\documentclass[runningheads]{llncs}

\usepackage{eccv}
\usepackage[width=122mm,left=12mm,paperwidth=146mm,height=193mm,top=12mm,paperheight=217mm]{geometry}

\usepackage{eccvabbrv}

\usepackage{graphicx}
\usepackage{booktabs}
\usepackage{enumitem}
\usepackage{booktabs}
\usepackage{pifont}
\usepackage{multirow}
\usepackage{subcaption}

\newcommand{\cmark}{\ding{51}}
\newcommand{\xmark}{\ding{55}}

\usepackage[table]{xcolor}

\usepackage[accsupp]{axessibility}  

\usepackage[pagebackref]{hyperref}

\usepackage{orcidlink}
\newcommand{\datasetname}{SP-TransientBench}
\newcommand{\datasetabbr}{STB}
\newcommand{\binwidth}{750 ps}
\newcommand{\histbins}{672}

\begin{document}

\title{\datasetname: A Real-Captured Single Photon Perception Benchmark} 

\titlerunning{Abbreviated paper title}

\author{Hongzhou Dong\inst{1} \and
Zili Zhang\inst{1} \and
Ziting Wen\inst{2} \and
Yiheng Qiang\inst{1} \and
Runrong Deng\inst{1} \and
Wenle Dong\inst{1} \and
Ziwen Jiang\inst{1} \and
Xinyang Li\inst{1} \and
Rui Lu\inst{1} \and
Shuoyao Sun\inst{1} \and
Wenyu Wang\inst{1} \and
Ziyi Xia\inst{1} \and
Haitao Zheng\inst{1} \and
Guodong Shi\inst{3} \and
Xiaoqiang Ren\inst{1}}

\authorrunning{H.~Dong et al.}

\institute{Shanghai University, Shanghai, China \and
Southern University of Science and Technology, Shenzhen, China \and
The University of Sydney, Sydney, Australia}

\maketitle

\begin{abstract}

Single-photon LiDAR (SPL) based on
single-photon avalanche diode (SPAD) sensing
enables time-resolved photon measurements with extreme sensitivity, offering unique potential for active 3D perception in photon-starved scenarios.
However, real-world single photon perception remains fundamentally challenging due to unique measurement noise and complex multi-return transient phenomena, which jointly complicate geometric reconstruction and semantic scene understanding. Despite growing interest in SPAD-based sensing, existing studies are largely limited to simulated data or small-scale controlled captures. As a result, systematic evaluation of real-world single photon perception across depth estimation, multi-view reconstruction, and 3D semantic understanding remains underexplored. To bridge this gap, we introduce \datasetname\ (\datasetabbr), a real-captured multi-task benchmark for single photon perception. \datasetabbr\ comprises 10 diverse scenes and 10,297 views captured using a solid-state single-photon LiDAR at $256\times192$ resolution. Each view provides full time-of-flight histograms with multi-return behavior, standardized metadata, and calibrated camera poses for multi-view evaluation. We further provide 13-class 3D semantic annotations for selected scenes. By providing dedicated data splits and evaluation protocols for each task, \datasetabbr\ enables consistent and reproducible benchmarking of real-world single photon perception across multiple 3D vision problems. 
The dataset and code will be released upon acceptance.

\end{abstract}

\begin{figure}[t]
\centering
\includegraphics[width=0.8\linewidth]{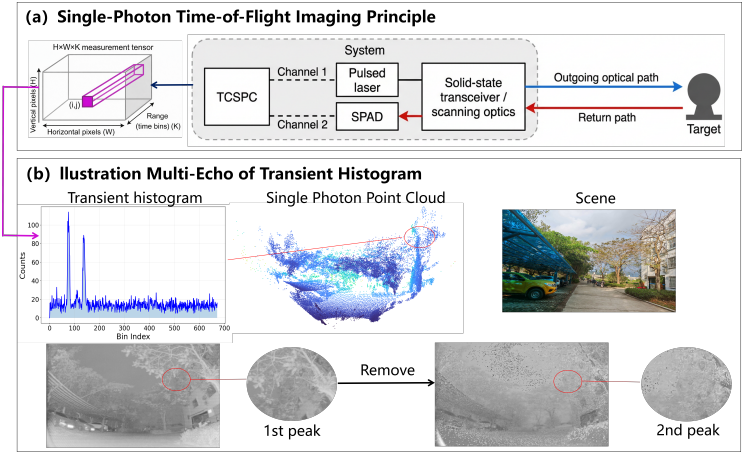}
\caption{Single-photon LiDAR (SPL) sensing pipeline and representative observations. (a) System overview of a SPAD-based SPL setup. A pulsed laser illuminates the scene and returned photons are detected by a SPAD and time-stamped via a TCSPC module, forming an $H \times W \times K$ transient measurement tensor.
(b) Example real-captured measurements and reconstructions. Transient histograms may exhibit multi-peak structures due to multi-path or multi-surface returns, which propagate to downstream point-cloud reconstruction and can cause noise, sparsity, and local structural ambiguities in challenging regions (highlighted).}
\label{fig:overview}
\end{figure}

\section{Introduction}
\label{sec:intro}

Single-photon LiDAR (SPL) has emerged as a powerful modality for active 3D perception due to its extreme photon sensitivity. 
By detecting and time-stamping individual photons, SPL systems operate under low-power illumination while achieving centimeter-level depth precision and kilometer-scale ranging~\cite{rapp2020advances}. 
Unlike conventional pulsed or amplitude-modulated LiDAR systems, which typically report a limited number of discrete echo returns per pixel, SPL records full time-resolved photon arrival histograms.
This capability enables explicit modeling of transient light transport, multi-surface returns, and fine temporal structures~\cite{2024Transientangelo, Malik2023TransientNeRF}. 
As illustrated in Fig.~\ref{fig:overview} (a), a pulsed laser illuminates the scene, and returned photons are captured by a Single Photon Avalanche Diode (SPAD) detector and time-stamped via a time-correlated single-photon counting (TCSPC) module, forming a three-dimensional transient tensor of size $H \times W \times K$. 
These properties make SPL particularly attractive for long-range perception in autonomous driving, remote sensing, and non-line-of-sight (NLOS) imaging~\cite{ovren2025long, young2025enhancing, scheuble2025transient}.

The sensing principle of SPL fundamentally differs from that of conventional imaging modalities. 
Photon detections are discrete stochastic events governed by Poisson statistics, and measurements are accumulated as time-of-flight histograms over repeated illuminations~\cite{shin2015single}. 
In practice, signal photons are sparse and often contaminated by ambient illumination, resulting in a low signal-to-background ratio (SBR)~\cite{altmann2016robust}.
Additional factors such as pulse broadening, attenuating media, detector dead time, and temporal quantization further distort transient measurements~\cite{tobin2019three,rapp2019dead}.
As a result, transient histograms frequently exhibit multi-peak structures arising from multi-path and multi-surface returns (Fig.~\ref{fig:overview} (b)), which propagate to downstream point-cloud reconstruction and induce noise, sparsity, and local geometric ambiguities. 
Robust single photon perception therefore requires both accurate physical modeling of photon statistics and strong structural priors over scene geometry.

Recent years have seen rapid progress in single photon perception, driven by advances in transient imaging models, signal processing techniques, and learning-based reconstruction frameworks~\cite{Malik2023TransientNeRF,time_resolved_mnist,20SPADnet,otoole2017transient}. 
While these methods demonstrate promising results on simulated data or small-scale controlled captures, the community still lacks a large-scale, real-captured benchmark that supports systematic multi-task evaluation. 
This gap is becoming increasingly limiting as SPL research transitions from methodological exploration toward real-world deployment.

Constructing such a benchmark presents two major challenges. 
First, SPAD-based SPL hardware remains specialized and relatively inaccessible, with customized acquisition pipelines that constrain scalable data collection and standardized release. 
Second, annotating transient measurements is substantially more complex than labeling RGB-D images or conventional point clouds. 
A single pixel histogram may contain multiple temporally separated peaks corresponding to distinct surfaces along a ray~\cite{shin2016computational}. 
These peaks can overlap or be obscured by background noise, making the mapping between temporal bins and physical surfaces inherently ambiguous. 
Unlike conventional labeling pipelines that operate on single-depth maps or consolidated point clouds, transient data require direct histogram-domain annotation and explicit handling of multi-return structures, which is not supported by existing tools and typically requires extensive manual disambiguation. 
Consequently, existing SPAD datasets remain limited in scale, realism, or task coverage.

To overcome these limitations, we introduce \datasetname\ (\datasetabbr), a real-captured multi-task SPL benchmark with full time-of-flight histograms, calibrated multi-view poses, and consistent 3D semantic annotations. Our main contributions are summarized as follows:\\
(1) We introduce \datasetname\ (\datasetabbr), the first publicly available real-captured multi-task SPL dataset for 3D perception with full time-of-flight histograms, calibrated multi-view poses, and 3D semantic annotations.\\
(2) We establish a standardized SPL benchmark covering depth estimation, multi-view reconstruction, and 3D semantic segmentation with unified splits and evaluation protocols.\\
(3) We develop a histogram-domain semantic annotation framework and release dedicated tools for labeling multi-return transient measurements.

\section{Related Work}
\label{sec:related}

\subsection{SPAD Datasets and Benchmarks}
\label{sec:related_datasets}

Time-resolved SPAD sensing has led to a growing number of datasets for single-photon imaging and transient-based 3D reconstruction. 
Existing releases can be broadly categorized into three groups: (i) real-captured transient imaging datasets, (ii) simulation-based SPL depth datasets, and (iii) multi-view transient rendering datasets.

Early real-captured datasets demonstrate waveform-based reconstruction from SPAD measurements~\cite{otoole2017transient,gutierrez2022compressive}, but primarily focus on single-view depth recovery and typically lack calibrated multi-view geometry or semantic annotations. 
Simulation-based datasets~\cite{20SPADnet,time_resolved_mnist,2023High} enable controlled modeling of photon statistics, yet may not fully reproduce real-world multi-return effects and sensor non-idealities. 
More recent works such as TransientNeRF~\cite{Malik2023TransientNeRF} combine real and simulated measurements for neural transient rendering.

Overall, existing datasets are designed for specific subtasks under heterogeneous acquisition protocols, limiting systematic cross-method evaluation. 
To date, no real-captured SPL benchmark jointly supports depth estimation, multi-view 3D reconstruction, and semantic perception under standardized splits and evaluation protocols. 
\datasetabbr\ addresses this gap by providing full time-resolved histograms together with calibrated multi-view poses within a unified real acquisition framework. 
Table~\ref{tab:spad_dataset_comparison} summarizes representative datasets for comparison.

\subsection{SPAD Processing, Reconstruction, and Perception}
\label{sec:related_processing}

SPL processing aims to recover scene depth and reflectivity from sparse photon detections affected by background noise and temporal response effects. 
Early approaches relied on histogram peak detection and waveform fitting~\cite{2000Laser,2016Waveform,2017A}, followed by photon-efficient and probabilistic formulations that improved robustness under low-flux conditions~\cite{kirmani2014first,shin2015photon,shin2016photon,shin2016computational,2016A,2017Restoration}.

More recently, learning-based methods directly map time-resolved histograms to depth or reflectivity~\cite{peng2020photon,yao2022robust,yao2022dynamic}. 
Time-resolved measurements have also been incorporated into neural implicit representations for multi-view reconstruction and transient rendering~\cite{Attal2021ToRF,Malik2023TransientNeRF,2024Transientangelo}, and further extended to semantic perception from SPL-derived depth maps or point clouds~\cite{esp_yolo,unet_spad,wen2026noiseaware}.

Despite rapid algorithmic progress, evaluation is typically conducted on proprietary captures or simulation-based datasets under heterogeneous protocols. 
\datasetabbr\ provides a standardized real-captured benchmark that enables systematic and fair evaluation across depth estimation, multi-view reconstruction, and 3D semantic understanding.

\begin{table*}[t]
\centering
\caption{Comparison of representative SPAD datasets and benchmarks.}
\label{tab:spad_dataset_comparison}
\resizebox{0.8\textwidth}{!}{%
\begin{tabular}{lcccccc}
\toprule
Dataset &
Annotations &
Pose &
Depth &
Sim &
Real &
Task Focus \\
\midrule
O'Toole et al.~\cite{otoole2017transient} &
\xmark &
\xmark & \xmark &
\textemdash
 & 7 &
Imaging \\

Bian et al.~\cite{2023High} &
\xmark &
\xmark & \xmark &
2{,}790 & \textemdash &
Imaging \\

SPADNet~\cite{20SPADnet} &
\xmark &
\xmark & \cmark &
8{,}266 & \textemdash
 &
Imaging \\

Time-Resolved MNIST~\cite{time_resolved_mnist} &
\cmark &
\xmark & \xmark &
70{,}000 & \textemdash
 &
Semantics \\

TransientNeRF~\cite{Malik2023TransientNeRF} &
\xmark &
\xmark & \cmark &
100 & 20 &
Reconstruction \\

\midrule
\textbf{\datasetabbr\ (Ours)} &
\cmark &
\cmark & \cmark &
\textemdash
 & 10{,}317  &
General-Purpose\\
\bottomrule
\end{tabular}%
}
\end{table*}

\section{Capture Platform}
\label{sec:sensor_platform}

To enable real-world data collection for single photon perception, we build a multi-sensor capture platform consisting of a solid-state single-photon LiDAR (Adaps ADS6311 Hawk~\cite{adaps_hawk}) and an auxiliary LiDAR (Livox Avia~\cite{livox_avia}). 
The Livox Avia is primarily used during dataset construction for pose estimation and depth reference generation.

\subsection{Hardware Overview}
\label{sec:sensor_hardware}

\noindent\textbf{Single-photon LiDAR.}
\datasetabbr\ is collected using an Adaps ADS6311 Hawk, a solid-state SPL device operating under a Direct Time-of-Flight (DToF) scheme with time-correlated single-photon counting (TCSPC). 
A VCSEL transmitter emits ultrashort laser pulses, and returned photons are detected by a synchronized SPAD array. 
Photon timestamps from repeated shots are accumulated into per-pixel time-of-flight histograms, from which range is estimated based on the emission–reception delay. The native SPAD resolution is $768\times576$. With $3\times3$ binning, the device outputs per-frame range measurements at $256\times192$. 
It provides a field of view of $128^\circ\times96^\circ$ (horizontal $\times$ vertical) and operates at 10--20\,Hz. 

\noindent\textbf{Auxiliary LiDAR.}
A Livox Avia LiDAR is mounted as an auxiliary sensor to support pose estimation and depth reference generation. 
The device outputs dense 3D point clouds and integrates an onboard IMU with timestamped measurements. 
In our pipeline, the point clouds are processed using mature LiDAR-inertial SLAM frameworks~\cite{kummerle2011g} to estimate drift-reduced sensor trajectories. 
These trajectories are transformed into the SPL coordinate frame via calibrated extrinsics and used as per-view camera poses, benefiting from the robustness and maturity of conventional LiDAR-based SLAM for geometric localization.

For the depth estimation track, Livox measurements are treated as reference ground truth. 
The point clouds are registered to the SPL coordinate system and projected into depth maps to provide metrically consistent supervision. 
Although LiDAR measurements are subject to ranging noise and cross-sensor calibration errors, they offer dense and geometrically reliable depth references for benchmarking SPL-based reconstruction methods.

\noindent\textbf{Ambient Illumination Measurement.}
We record ambient illumination intensity during data capture using a calibrated light meter. 
These measurements are used for dataset statistics and analysis of sensing conditions.







In SPAD-based SPL sensing, photon detections are timestamped relative to the emission clock and discretized into uniform temporal bins. 
Accumulating timestamps over repeated laser shots yields a per-pixel time-of-flight histogram. 
The temporal location of a peak encodes the round-trip travel time and thus scene depth, while its magnitude reflects return intensity; background illumination contributes an approximately uniform noise floor. 
Multi-surface or multi-path propagation may produce multiple peaks within a single histogram, motivating echo-level interpretation.

In \datasetabbr, each histogram consists of $T=\histbins$ bins with temporal resolution $\Delta t=\binwidth$. 
We release full time-resolved waveforms rather than only peak-based depth estimates, enabling explicit modeling of multi-return transients and background statistics. 
This design supports algorithms that reason over peak structure, background contamination, and range-dependent sparsity directly in the histogram domain.

\begin{figure}[t]
\centering
\begin{subfigure}[t]{0.45\linewidth}
    \centering
    \includegraphics[width=\linewidth]{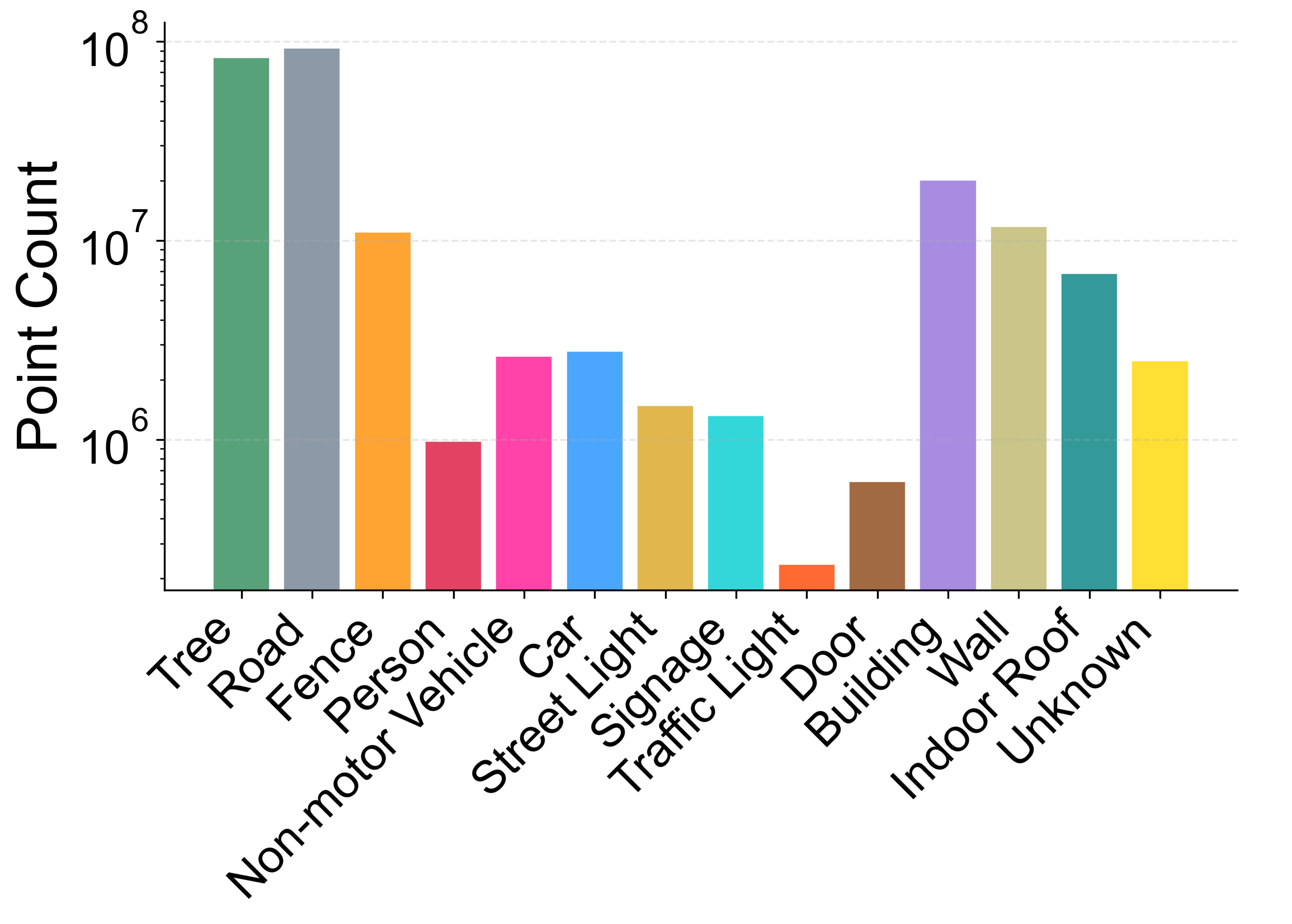}
    \caption{Distribution of semantic classes}
    \label{fig:class_distribution}
\end{subfigure}
\hspace{0.02\linewidth}
\begin{subfigure}[t]{0.45\linewidth}
    \centering
    \includegraphics[width=\linewidth]{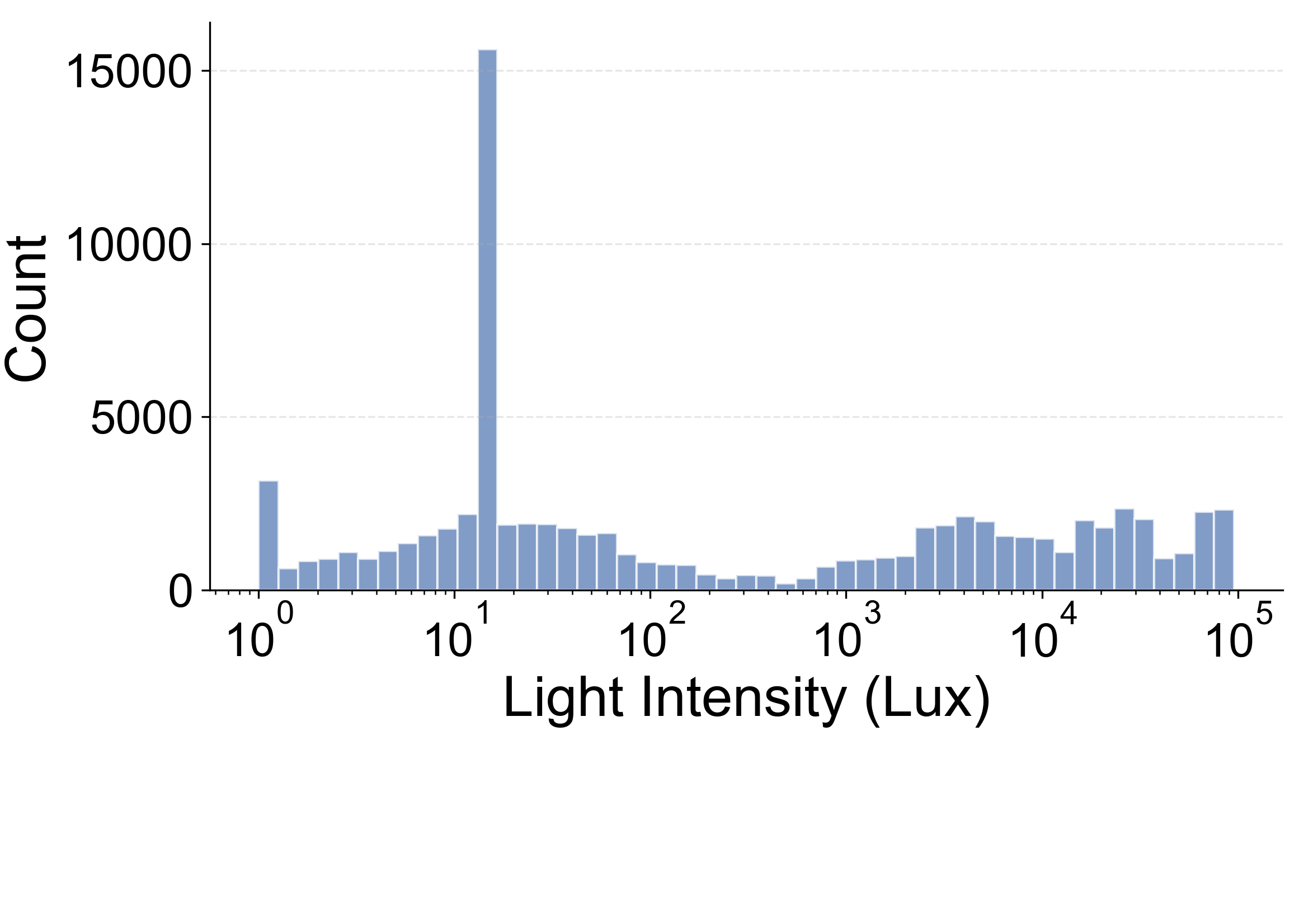}
    \caption{Distribution of light intensity}
    \label{fig:light_intensity_distribution}
\end{subfigure}
\caption{Statistical overview of semantic and illumination distributions in \datasetabbr.}
\label{fig:dataset_overview}
\end{figure}

\section{Dataset}
\label{sec:dataset}
\subsection{Overview and Statistics}
\label{sec:dataset_overview}

\begin{figure}[t]
\centering
\includegraphics[width=0.8\linewidth]{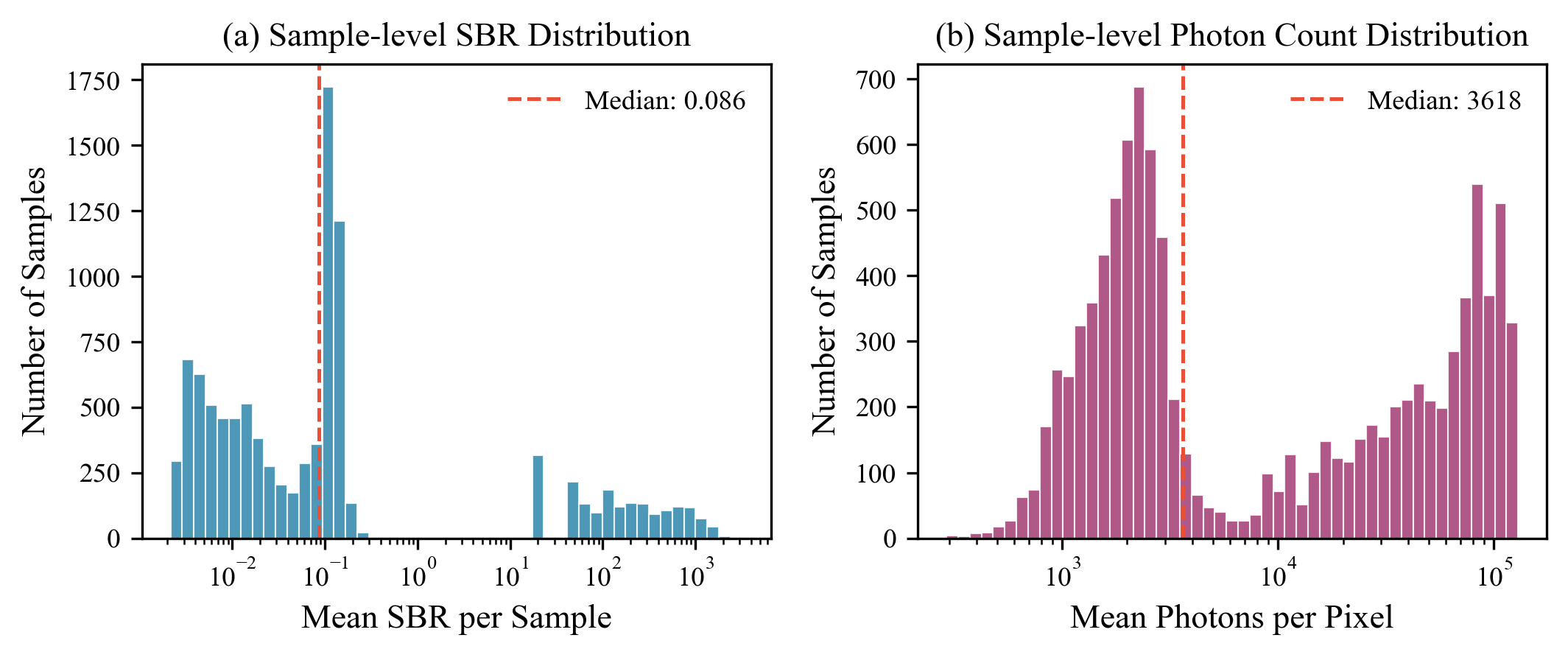}
\caption{Key dataset statistics, including range distribution and measurement factors: Signal-to-Background Ratio (SBR) and Mean Photons Per Pixel (MPPP).}
\label{fig:stats_range_sbr}
\end{figure}

We introduce \datasetname\ (\datasetabbr), a real-captured benchmark for single photon perception. 
The current release comprises 10 scenes and 10{,}297 views spanning diverse indoor and outdoor environments. 
Each view provides full time-of-flight histograms together with standardized metadata and annotations supporting three benchmark tracks: 
(i) depth estimation, 
(ii) multi-view 3D reconstruction, and 
(iii) 3D semantic segmentation. 
Additionally, each capture is associated with measured ambient illumination intensity, enabling analysis of dataset distribution under varying lighting conditions.
Figure~\ref{fig:dataset_overview} illustrates the distribution of semantic categories and recorded ambient illumination levels. 
Figure~\ref{fig:stats_range_sbr} further presents key sensing statistics, including range distribution and photon-limited measurement factors such as signal-to-background ratio (SBR) and mean photons per pixel (MPPP). 
Overall, \datasetabbr\ spans a wide spectrum of photon regimes, from higher-SBR, denser-return conditions to strongly background-dominated, sparse-return scenarios. 
This diversity is essential for evaluating the robustness and generalization of transient-based depth recovery and downstream 3D perception models under realistic SPL operating conditions.



\subsection{Dataset Organization}
\label{sec:dataset_structure}

Each scene in \datasetabbr\ is organized as a collection of views captured under a unified SPL configuration. 
The only shared core data across all tasks is the raw time-resolved histogram of size $256\times192\times672$ (full waveform) for each view.

\noindent\textbf{Geometric calibration.}
Per-view camera poses are estimated using an auxiliary Livox Avia LiDAR via LiDAR SLAM~\cite{ICP,kummerle2011g}.
The estimated trajectories are transformed into the SPL coordinate frame through a rigid extrinsic calibration obtained via checkerboard-based alignment between the Livox and SPL sensors.
SPL intrinsic parameters, including focal length, principal point, and distortion coefficients, are calibrated using a checkerboard procedure with the MATLAB Camera Calibration Toolbox, enabling consistent geometric back-projection of histogram-derived depth.

\noindent\textbf{Instrument response characterization.}
The instrument response function (IRF) of the SPL system is measured by imaging a high-reflectance ($99\%$) Enhanced Specular Reflector (ESR) target at a fixed range. 
The fitted temporal profile characterizes pulse broadening and temporal resolution, and is released together with the dataset. 
Detailed calibration procedures are provided in the supplementary material.

\noindent\textbf{Depth reference generation.}
For the depth estimation track, we provide a LiDAR-based reference depth when available. 
Auxiliary Livox point clouds are temporally accumulated to improve geometric stability and projected into the SPL viewpoint using calibrated poses to obtain metrically consistent reference depth maps.

\noindent\textbf{Task-specific releases.}
In addition to the shared SPL histograms, each benchmark track provides different auxiliary data depending on its evaluation requirements. 
Table~\ref{tab:track_release} summarizes the released components for each track.
For the semantic understanding track, 3D semantic annotations are generated using a dedicated histogram-domain labeling framework tailored to multi-return transient measurements. 
This annotation methodology enables consistent assignment of multiple semantic labels along a single pixel ray. 
Details of the annotation pipeline are provided in Sec.~\ref{sec:dataset_semantic}.

\begin{table}[t]
\centering
\caption{Per-track release contents in \datasetabbr. All tracks include the raw SPAD histogram of size $256\times192\times672$.}
\label{tab:track_release}
\setlength{\tabcolsep}{6pt}
\small
\resizebox{\linewidth}{!}{%
\begin{tabular}{lcccccc}
\toprule
Track & SPAD histogram & Livox data & SPAD--Livox extrinsics & Pose & Labels & Light intensity \\
\midrule
Depth Estimation              & \cmark & \cmark & \cmark & \xmark & \xmark & \xmark \\
Multi-view Reconstruction  & \cmark & \cmark & \cmark & \cmark & \xmark & \xmark\\
Semantic Understanding     & \cmark & \xmark & \xmark & \xmark & \cmark & \cmark \\
\bottomrule
\end{tabular}%
}
\end{table}

\subsection{3D Semantic Annotations}
\label{sec:dataset_semantic}
To support the semantic understanding track described above, we design a dedicated annotation pipeline tailored to multi-return SPL measurements.

\begin{figure}[t]
\centering
\IfFileExists{fig/annotation_pipeline.png}{
  \includegraphics[width=0.8\linewidth]{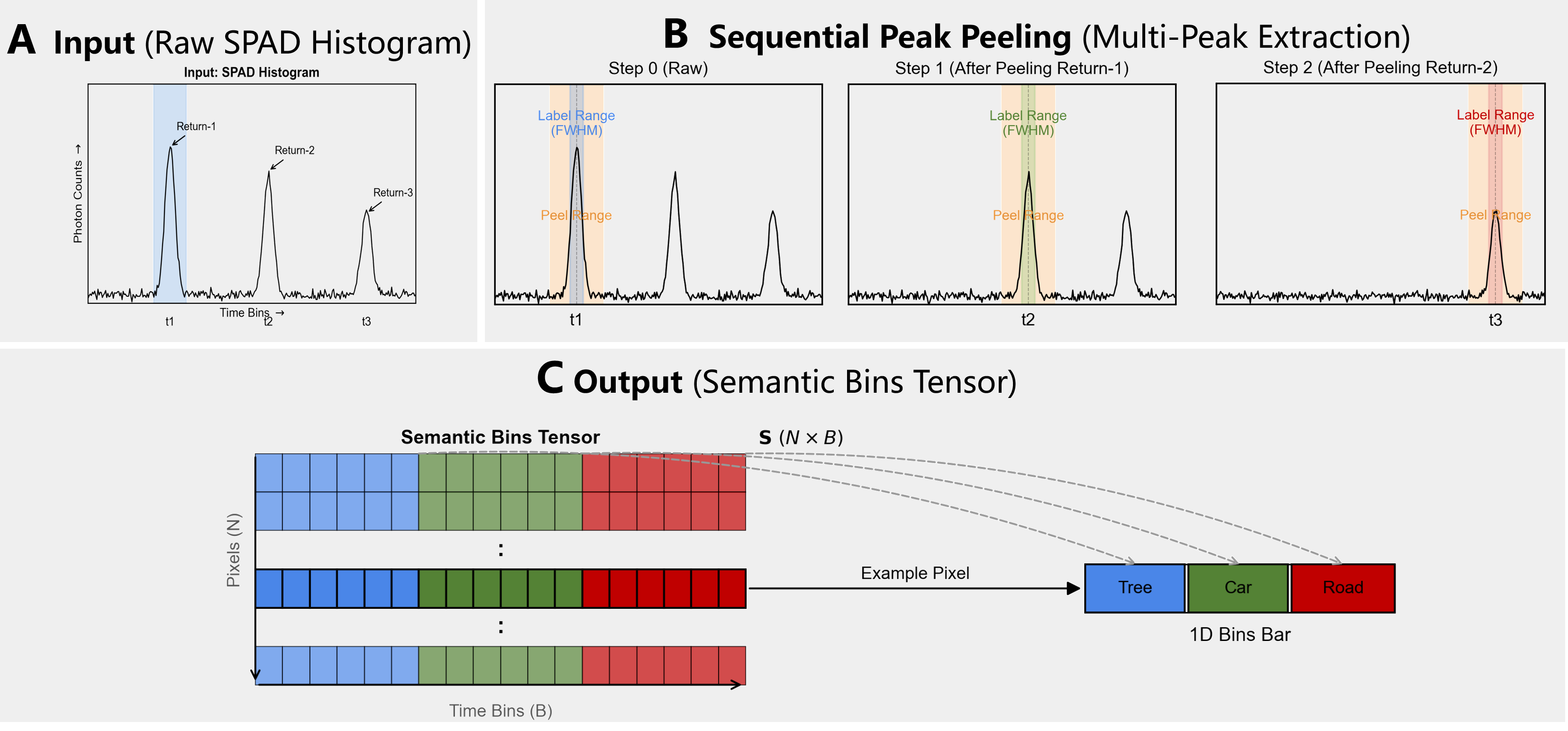}
}{
  \fbox{\parbox[c][3.2cm][c]{0.95\linewidth}{\centering
  Annotation pipeline placeholder (fig/annotation_pipeline.png).}}
}
\caption{Sequential peak labeling for multi-return semantic bin annotations. We label bin segments in the histogram domain and peel local peak support to reveal subsequent returns.}
\label{fig:annotation_pipeline}
\end{figure}

\noindent\textbf{Histogram-Domain 3D Semantic Annotation via Sequential Peak Peeling.}
To leverage the multi-return nature of SPAD measurements, we move beyond conventional single-depth labeling and introduce a dense annotation scheme defined directly in the histogram domain. 
As illustrated in Fig.~\ref{fig:annotation_pipeline}, the ground-truth labels are represented as a semantic bin tensor $\mathbf{S} \in \{0, \dots, C\}^{N \times B}$, where $N = H \times W$ denotes the number of pixels and $B$ the number of temporal bins.

We adopt an iterative sequential peak peeling strategy to resolve overlapping and multi-peak signals (Fig.~\ref{fig:annotation_pipeline}-B). 
Specifically: 
(1) \textbf{Peak Identification}: the dominant peak (e.g., return at $t_1$) is detected in the raw histogram; 
(2) \textbf{Interval Assignment}: a semantic label is assigned to the peak-support interval, defined using the full width at half maximum (FWHM) to maintain temporal precision; 
(3) \textbf{Signal Peeling}: the signal within a broader neighborhood of the labeled peak is suppressed to expose subsequent, weaker returns (e.g., $t_2, t_3$). 
This procedure is repeated until all valid returns are annotated.

The resulting tensor $\mathbf{S}$ enables a single pixel ray to encode multiple disjoint semantic entities (e.g., a semi-transparent Tree canopy followed by Road and Car, see Fig.~\ref{fig:annotation_pipeline}-C). 
For geometric supervision, each labeled bin is converted from temporal index to metric range and back-projected to 3D using calibrated intrinsics, producing dense volumetric semantic supervision.

To support this process, we develop an annotation tool tailored to time-of-flight histograms with a multi-return structure. 
The tool is used to generate all semantic bin annotations in \datasetabbr\ and will be released together with the dataset and evaluation code.

\section{Experiments}
\label{sec:benchmark}

We evaluate algorithms using {SPAD-only inputs}, without RGB guidance or RGB--SPAD fusion.
The benchmark includes three task families:
(i) {depth estimation},
(ii) {multi-view 3D reconstruction and novel-view evaluation},
and (iii) {3D semantic segmentation} from SPAD-derived geometry.

\subsection{Task 1: Depth Estimation}
\label{sec:task_depth}

\noindent\textbf{Baselines} To evaluate depth recovery from time-resolved transient histograms, we benchmark four representative methods: Shin~\cite{shin2016photon}, Rapp~\cite{2016A}, Li~\cite{li2020single}, and SSPINET~\cite{yao2022dynamic}. 
All methods estimate depth directly from raw photon time-of-flight histograms.
These baselines provide broad coverage of prevailing approaches in histogram-based depth estimation. 
Shin et al.~\cite{shin2016photon} formulate photon arrivals using a probabilistic image-formation model and solve a regularized optimization problem, enabling stable reconstruction in photon-starved regimes. 
Rapp et al.~\cite{2016A} propose a statistical signal–background unmixing framework that explicitly models ambient illumination and noise. 
Li et al.~\cite{li2020single} focus on long-range, extremely low-return scenarios with photon-efficient reconstruction. 
SSPINET~\cite{yao2022dynamic} represents learning-based approaches, leveraging sparsity-aware neural priors for transient-to-depth inference.
Together, these methods encompass model-based optimization, statistical inference, photon-efficient reconstruction, and data-driven learning, forming a comprehensive evaluation suite for assessing performance and generalization on STB.


\noindent\textbf{Evaluation Metrics} 
Following the protocol of Scheuble et al.~\cite{Scheuble_2025_ICCV}, we evaluate depth reconstruction using Chamfer Distance (CD) and Recall. 
CD measures geometric discrepancy between predicted and reference point clouds after back-projecting depths, while Recall measures the fraction of ground-truth points recovered within a distance threshold. 
The thresholds are defined by the temporal resolution of the transient measurements (750\,ps per bin). 
We report Recall at tolerances of 1, 3, and 5 temporal bins to analyze reconstruction accuracy under different geometric tolerances.

\begin{table}[t]
\centering
\caption{\textbf{Depth Estimation on \datasetabbr.} 
Chamfer Distance (CD, m) and Recall evaluated at different temporal tolerances.
}

\label{tab:imaging}
\setlength{\tabcolsep}{5pt}
\small
\begin{tabular}{ccccc}
\toprule
\multirow{2}{*}{Method} 
& \multirow{2}{*}{CD $\downarrow$ (m)} 
& \multicolumn{3}{c}{Recall $\uparrow$ (\%)} \\
\cmidrule(lr){3-5}
& & 1bin & 3bins & 5bins \\
\midrule
Shin~\cite{shin2016photon}  & 1.6908 & 25.97 & 74.83 & 89.14 \\
Rapp~\cite{2016A}          & 1.7486 & 29.17 & 68.62 & 83.88 \\
Li~\cite{li2020single}     & 1.5506 & 34.60 & 77.20 & 88.60\\
SSPINET~\cite{yao2022dynamic} & 1.4978 & 44.27 & 83.72 & 93.68 \\
\bottomrule
\end{tabular}
\end{table}

\begin{figure}[t]
\centering
\includegraphics[width=0.9\linewidth]{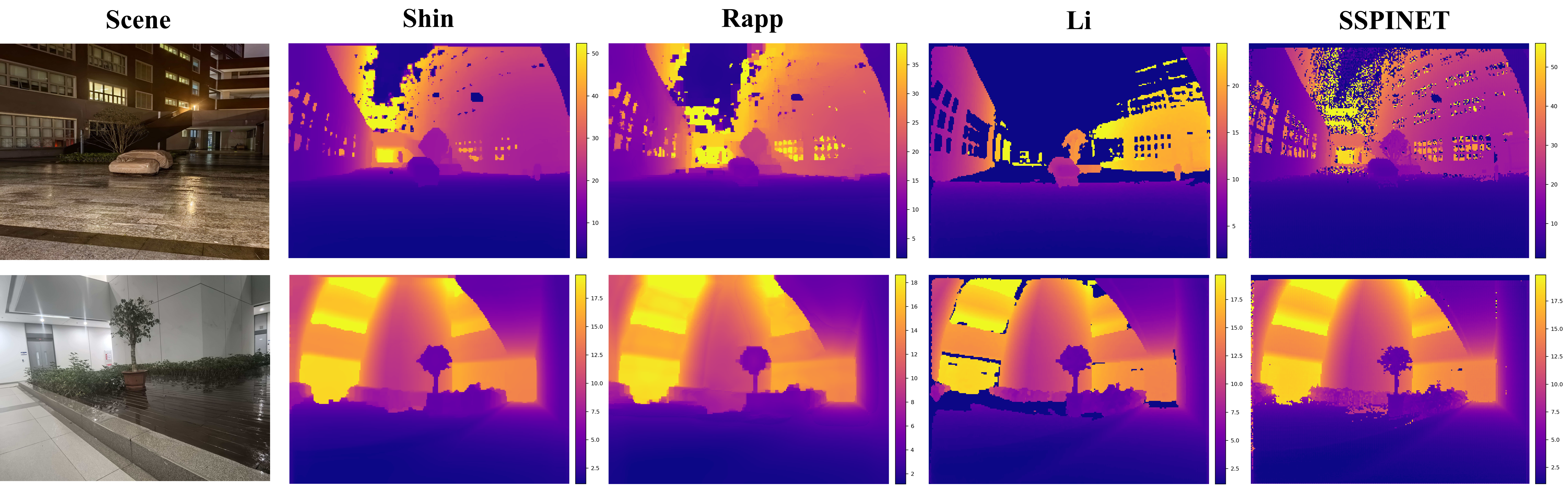}
\caption{Qualitative results for depth estimation on \datasetabbr.
Columns correspond to ground truth, Shin~\cite{shin2016photon}, Rapp~\cite{2016A}, Li~\cite{li2020single}, and SSPINET~\cite{yao2022dynamic}.}
\label{fig:depth_estimation}
\end{figure}

\noindent\textbf{Results and Discussion} Table~\ref{tab:imaging} summarizes depth reconstruction performance on \datasetabbr. 
SSPINET achieves the lowest CD and the highest Recall across all tolerance levels, while model-based methods remain competitive under relaxed thresholds but show reduced Recall at stricter tolerances (1 bin).

The gap between Recall (1 bin) and Recall (5 bins) highlights the effect of temporal tolerance on reconstruction accuracy. 
While CD values are relatively close across methods, Recall varies more noticeably, indicating differences in reconstruction completeness. Figure~\ref{fig:depth_estimation} shows representative qualitative results. Additional visualizations are provided in the supplementary material.

\subsection{Task 2: Multi-View 3D Reconstruction}
\label{sec:task_recon}
\noindent\textbf{Baselines} \label{sec:recon_baselines}
We focus on the multi-view 3D reconstruction track, which aims to recover scene geometry from multiple SPL views using camera poses for view alignment. 
To evaluate \datasetabbr\ on this task, we adopt representative baselines including TransientNeRF~\cite{Malik2023TransientNeRF}, Transientangelo~\cite{2024Transientangelo}.
These methods represent typical transient-based rendering approaches that reconstruct scene geometry directly from single-photon time-of-flight histograms.



\noindent\textbf{Implementation details}
Models were trained on a single RTX 4090 using Adam with an initial learning rate of $1\times10^{-3}$ for 100k iterations with a batch size of 512 pixels. For each scene, we construct training sets using 3, 5, or 10 views, respectively, while reserving the remaining views for novel-view rendering and geometry evaluation. 
Unless otherwise specified, results reported in the main paper correspond to the 10-view setting. 
Results under the 3-view and 5-view configurations are included in the supplementary material for completeness.

\noindent\textbf{Evaluation Metrics} Following the previous work~\cite{scheuble2025transient}, we assess reconstruction quality under three output domains, depending on the predicted representation and the rendered modality at novel views.
\textbf{(i) Intensity rendering:} We report Structural Similarity Index Measure (SSIM) and Learned Perceptual Image Patch Similarity (LPIPS) between rendered intensity images and references.
\textbf{(ii) Depth rendering:} We compute the per-pixel $L_1$ error between rendered and reference depth maps over valid pixels.
\textbf{(iii) Histogram rendering:} For methods that render per-pixel transient histograms, we report PSNR (Peak Signal-to-Noise Ratio) in the histogram domain.


\begin{table}[t]
\centering
\caption{\textbf{Multi-view reconstruction on \datasetabbr.} 
We report novel-view rendering metrics for intensity (SSIM, LPIPS), depth ($L_1$), and histogram (PSNR) outputs. }
\label{tab:3d_recon}
\setlength{\tabcolsep}{5pt}
\small
\begin{tabular}{ccccc}
\toprule
\multirow{2}{*}{Method} & \multicolumn{2}{c}{Intensity} & \multicolumn{1}{c}{Depth} & \multicolumn{1}{c}{Histogram} \\
\cmidrule(lr){2-3}\cmidrule(lr){4-4}\cmidrule(lr){5-5}
& SSIM $\uparrow$ & LPIPS $\downarrow$ & $L_1$ $\downarrow$ & PSNR $\uparrow$ \\
\midrule
TransientNeRF~\cite{Malik2023TransientNeRF} 
& 0.7717
& 0.2315
& 0.7836
& 45.4251 \\

Transientangelo~\cite{2024Transientangelo}     
& 0.7597 & 0.2501 & 0.8710 & 41.5735 \\
\bottomrule
\end{tabular}
\end{table}

\begin{figure}[t]
\centering
\includegraphics[width=0.9\linewidth]{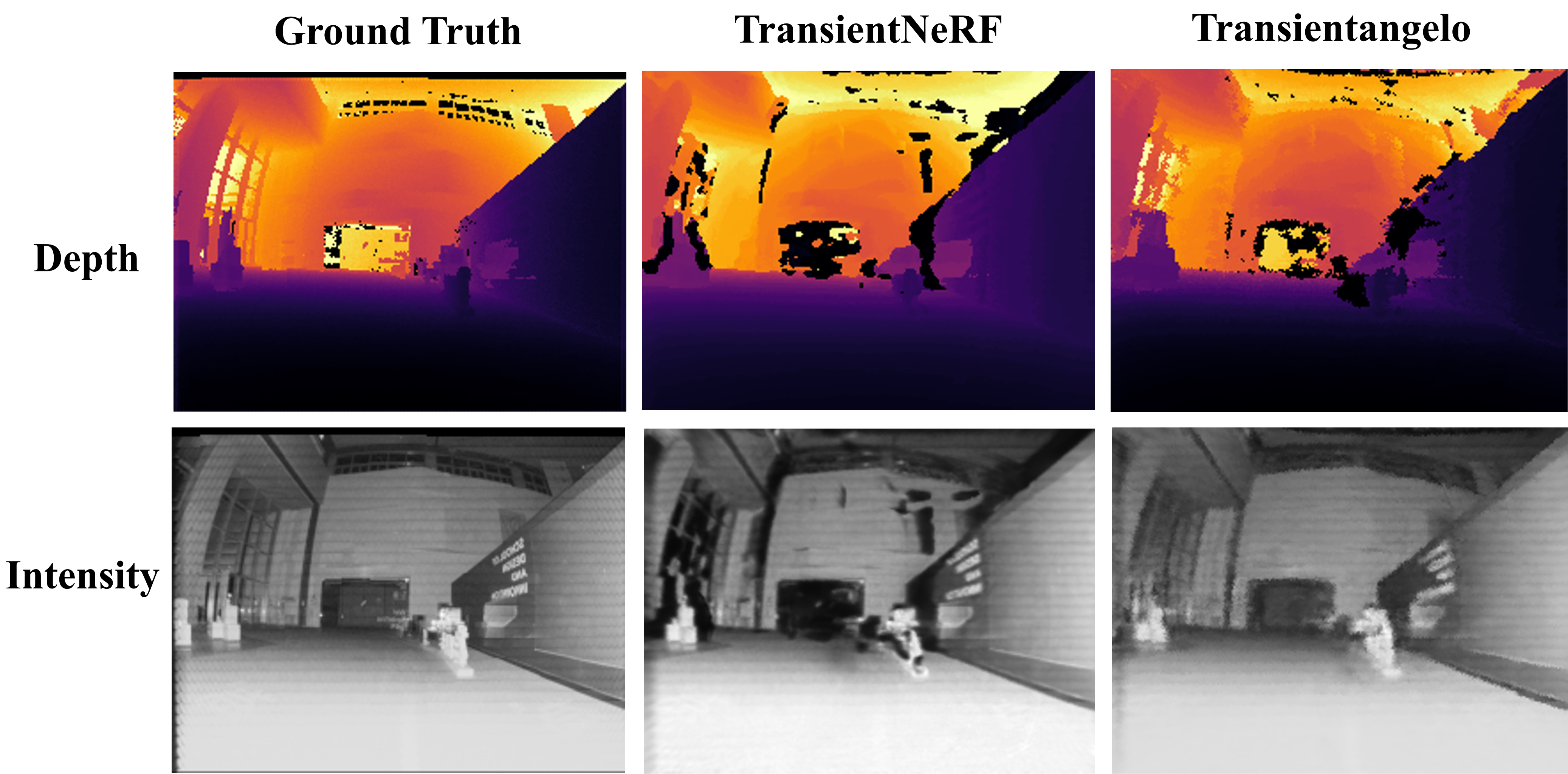}
\caption{Qualitative results for multi-view reconstruction on \datasetabbr.
Depth (top) and intensity (bottom) renderings at a novel view. Columns correspond to Ground Truth, TransientNeRF, Transientangelo.}
\label{fig:Multiview_reconstruction}
\end{figure}

\noindent\textbf{Results and Discussion}
Table~\ref{tab:3d_recon} reports quantitative results for multi-view reconstruction on \datasetabbr. 
We focus on representative approaches that reconstruct scene geometry and synthesize novel views directly from single-photon transient histograms. 
These results demonstrate the feasibility of recovering geometry and appearance from time-resolved SPL measurements.

Figure~\ref{fig:Multiview_reconstruction} presents representative qualitative examples of depth and intensity renderings at novel viewpoints. 
Additional analyses and comparisons with classical reconstruction pipelines that operate on depth or intensity images are provided in the supplementary material.

\subsection{Task 3: 3D Semantic Segmentation}
\label{sec:task_semantic}
\noindent\textbf{Baselines}
\label{sec:rep_baselines_sem}
We introduce the semantic understanding track of \datasetabbr, which targets 3D semantic segmentation from SPAD time-resolved measurements. To evaluate \datasetabbr\ under this setting, we benchmark representative pipelines that combine histogram-domain preprocessing with point-cloud segmentation backbones.

Since SPAD transient histograms are typically noisy along the temporal axis, we first apply preprocessing to enhance signal returns and then convert the processed histograms into single-photon point clouds through a fixed histogram-to-range projection. We consider four representative preprocessing methods: Thresholding, Matched Filtering~\cite{turin1960matched}, PPC~\cite{Goyal_2025_ICCV}, and SSPINET~\cite{yao2022dynamic}. Matched filtering is a classical peak-enhancement technique, while PPC and SSPINET represent SPAD-oriented photon processing and learning-based reconstruction approaches.

For point-cloud segmentation, we adopt four widely used backbones: PointNet++~\cite{qi2017pointnet++}, PointMLP~\cite{ma2022rethinking}, Point Transformer~\cite{zhao2021point}, and PointNeXt~\cite{qian2022pointnext}. As summarized in Tab.~\ref{tab:seg3d_preprocess_backbone}, we evaluate all preprocessing–backbone combinations to analyze the impact of histogram filtering and network design on semantic segmentation performance.

\noindent\textbf{Implementation details}
Models were trained on a single RTX 4090 using 8,297 training samples (2,000 test) for 100 epochs with batch size 64 and an initial learning rate of $1\times10^{-3}$; results are averaged over three random seeds.

\noindent\textbf{Evaluation Metrics}
\label{sec:eval_metrics_sem} Consistent with prevailing 3D semantic segmentation benchmarks, our evaluation framework is designed for a thorough assessment of semantic understanding performance on \datasetabbr. In particular, we focus on Overall Accuracy (OA) and mean Intersection-over-Union (mIoU) as the primary evaluation metrics, following common practice in point-cloud scene understanding. All reported results are the mean ± standard deviation computed across three independent runs using random seeds {1, 456, 789}.

\begin{figure}[t]
\centering
\includegraphics[width=0.9\linewidth]{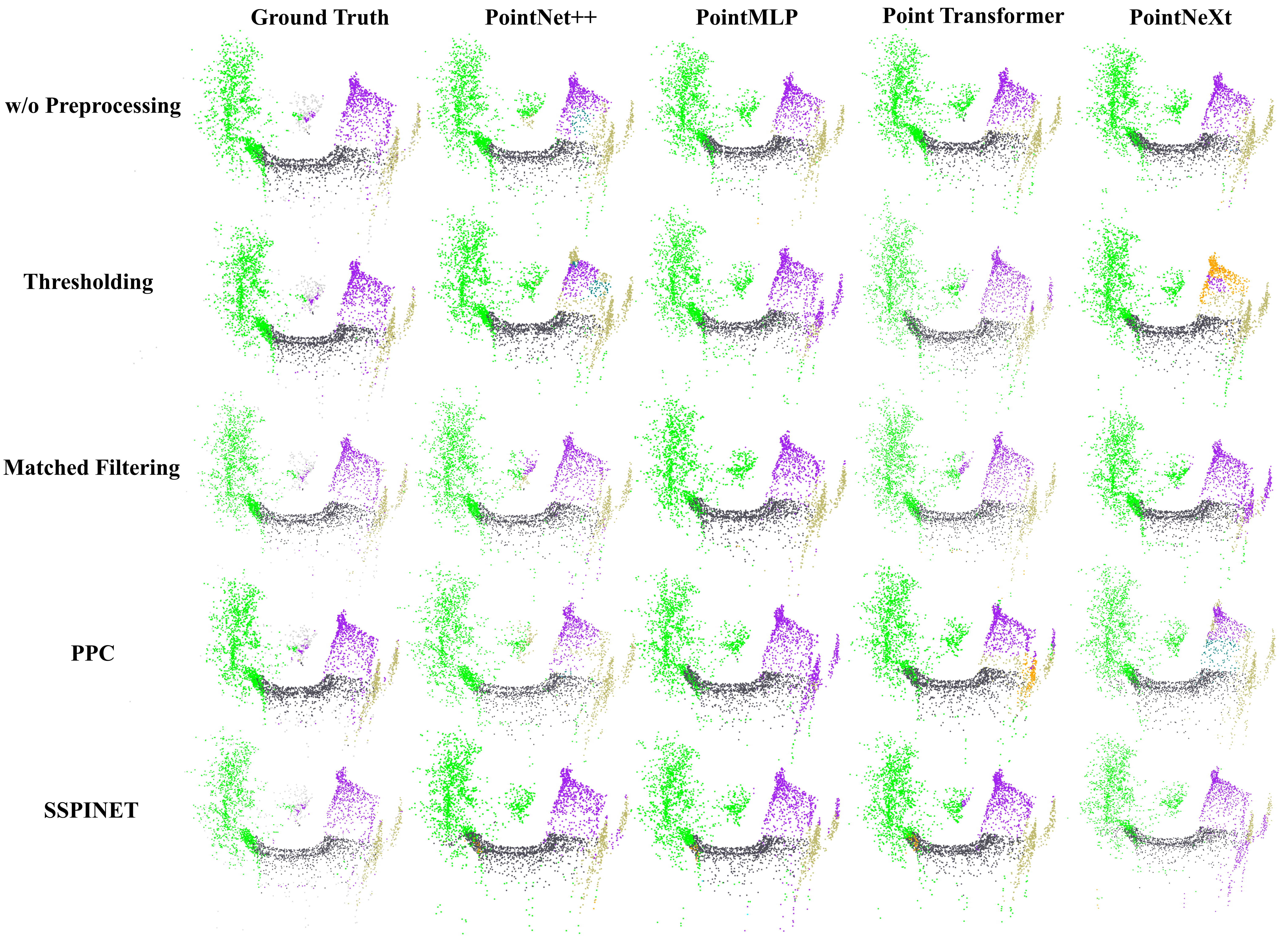}
\caption{Qualitative results for 3D semantic segmentation on \datasetabbr. 
Columns correspond to Ground Truth and predictions from different preprocessing and backbone combinations.}
\label{fig:semantic_segmentation}
\end{figure}

\noindent\textbf{Results and Discussion}
Table~\ref{tab:seg3d_preprocess_backbone} reports 3D semantic segmentation performance under different combinations of histogram-domain preprocessing and point-cloud backbones. Applying preprocessing consistently improves results over raw transient inputs. Even simple thresholding yields noticeable gains, indicating that suppressing background noise before histogram-to-point projection benefits downstream semantic learning. Learning-based preprocessing further provides stable improvements across most backbones. Across architectures, performance differences are relatively modest compared to the impact of preprocessing. Despite high OA, mIoU remains around 50\%, reflecting geometric sparsity and photon noise inherent to SPL measurements. Figure~\ref{fig:semantic_segmentation} shows representative qualitative segmentation results. Additional visualizations are provided in the supplementary material.

\begin{table*}[t]
\centering
\caption{3D semantic segmentation from SPAD transient inputs on \datasetabbr.} 
\label{tab:seg3d_preprocess_backbone}
\setlength{\tabcolsep}{5pt}
\small
\resizebox{\textwidth}{!}{%
\begin{tabular}{c|cc|cc|cc|cc}
\toprule
\multirow{2}{*}{Preprocessing} &
\multicolumn{2}{c|}{PointNet++~\cite{qi2017pointnet++}} &
\multicolumn{2}{c|}{PointMLP~\cite{ma2022rethinking}} &
\multicolumn{2}{c|}{Point Transformer~\cite{zhao2021point}} &
\multicolumn{2}{c}{PointNeXt~\cite{qian2022pointnext}} \\
\cline{2-9}
& OA $\uparrow$ & mIoU $\uparrow$
& OA $\uparrow$ & mIoU $\uparrow$
& OA $\uparrow$ & mIoU $\uparrow$
& OA $\uparrow$ & mIoU $\uparrow$ \\
\midrule
w/o Preprocessing 
& \(89.95 \pm 0.37\) & \(49.16 \pm 1.31\) 
& \(89.90 \pm 0.36\) & \(49.11 \pm 2.40\) 
& \(90.23 \pm 0.11\) & \(49.51 \pm 0.47\) 
& \(88.45 \pm 0.19\) & \(41.05 \pm 0.75\) \\

Thresholding 
& \(90.53 \pm 0.21\) & \(50.58 \pm 0.59\) 
& \(90.81 \pm 0.11\) & \(52.22 \pm 0.47\) 
& \(89.96 \pm 0.88\) & \(48.59 \pm 2.68\) 
& \(89.05 \pm 0.50\) & \(42.86 \pm 1.31\) \\

Matched Filtering~\cite{turin1960matched} 
& \(90.26 \pm 0.21\) & \(49.71 \pm 1.77\) 
& \(89.93 \pm 0.25\) & \(50.02 \pm 0.46\) 
& \(90.43 \pm 0.14\) & \(50.37 \pm 0.30\) 
& \(87.07 \pm 2.29\) & \(40.04 \pm 1.40\) \\

PPC~\cite{Goyal_2025_ICCV} 
& \(90.50 \pm 0.33\) & \(50.41 \pm 1.29\) 
& \(90.59 \pm 0.30\) & \(51.50 \pm 0.13\) 
& \(90.35 \pm 0.44\) & \(48.95 \pm 1.62\) 
& \(88.43 \pm 0.29\) & \(41.26 \pm 1.10\) \\

SSPINET~\cite{yao2022dynamic} 
& \(90.73 \pm 0.13\) & \(51.31 \pm 0.31\) 
& \(90.65 \pm 0.10\) & \(51.78 \pm 0.70\) 
& \(90.51 \pm 0.36\) & \(50.03 \pm 0.95\) 
& \(88.46 \pm 0.89\) & \(41.54 \pm 1.52\) \\

\bottomrule
\end{tabular}%
}
\end{table*}




\subsection{Simulation-to-Real Gap}
\label{sec:sim2real}

Most learning-based methods for single-photon imaging rely on simulated data due to the difficulty of collecting large-scale real SPL measurements with annotations~\cite{Scheuble_2025_ICCV}. 
However, simulated transients often fail to reproduce the complex noise statistics, multi-return effects, and sensor artifacts present in real observations, leading to a performance gap when models are applied to real SPL data.

\begin{table}[t]
\centering
\caption{Effect of real data ratio and pretraining strategy for 3D semantic segmentation.
Models are evaluated on real SPL measurements from \datasetabbr.}
\label{tab:sim2real}

\setlength{\tabcolsep}{2pt}
\renewcommand{\arraystretch}{1.05}
\scriptsize

\resizebox{\columnwidth}{!}{%
\begin{tabular}{lcccccccc}
\toprule
& \multicolumn{2}{c}{10\%} & \multicolumn{2}{c}{20\%} & \multicolumn{2}{c}{50\%} & \multicolumn{2}{c}{100\%} \\
\cmidrule(lr){2-3} \cmidrule(lr){4-5} \cmidrule(lr){6-7} \cmidrule(lr){8-9}
Training & OA & mIoU & OA & mIoU & OA & mIoU & OA & mIoU \\
\midrule
Scratch
& 83.80$\pm$0.79 & 33.19$\pm$1.37
& 86.16$\pm$0.68 & 37.98$\pm$1.09
& 89.09$\pm$0.36 & 45.94$\pm$1.37
& 91.02$\pm$0.11 & 52.24$\pm$0.36 \\
Finetune
& 85.58$\pm$0.34 & 36.57$\pm$0.61
& 87.52$\pm$0.25 & 41.39$\pm$1.42
& 89.55$\pm$0.56 & 47.98$\pm$0.93
& 90.61$\pm$0.81 & 51.20$\pm$1.41 \\
\bottomrule
\end{tabular}%
}
\end{table}

To study this gap, we conduct a controlled experiment on 3D semantic segmentation using PointNet++~\cite{qi2017pointnet++}. 
Simulated single-photon point clouds are generated from ScanNet~\cite{dai2017scannet} by converting depth maps into transient histograms with a photon noise model. 
We compare two training strategies: (i) \textbf{Scratch}, trained only on real data, and (ii) \textbf{Finetune}, initialized from simulation pretraining and further optimized on real captures.

Table~\ref{tab:sim2real} reports results using different fractions of the real training set. 
Simulation pretraining improves performance when only limited real data is available, indicating that simulated data provides useful feature initialization. 
However, this advantage diminishes as more real data is introduced. With the full training set, models trained from scratch achieve comparable or slightly better performance.

These results suggest that while simulated transient data offers useful priors, it cannot fully capture the characteristics of real SPL observations. 
Therefore, real-captured datasets such as \datasetabbr\ remain essential for developing robust single-photon perception algorithms.

\section{Conclusion}
\label{sec:conclusion}


We presented \datasetname\ (\datasetabbr), a real-captured benchmark for single-photon LiDAR (SPL) that enables systematic evaluation of time-resolved transient sensing for 3D perception. The dataset provides full per-pixel photon time-of-flight histograms, calibrated multi-view poses, and consistent 3D semantic annotations collected under a unified acquisition setup. By releasing raw transient measurements rather than only derived depth maps, \datasetabbr\ supports algorithmic development that directly models photon arrival statistics, multi-return structures, and background noise characteristics inherent to SPL sensing. Beyond the dataset itself, \datasetabbr\ establishes standardized tasks, official splits, and evaluation protocols covering depth estimation, multi-view 3D reconstruction, and 3D semantic understanding. These benchmarks enable fair and reproducible comparisons across diverse approaches, including signal-processing methods, physics-based reconstruction techniques, and learning-based perception models. To further facilitate research on transient data, we introduce a histogram-domain semantic annotation framework that supports scalable labeling of multi-return measurements. By providing both data and tooling tailored to SPL observations, we hope \datasetabbr\ will serve as a practical foundation for future studies on photon-limited 3D perception and transient-based scene understanding.

\noindent\textbf{Challenges and Outlook.}
STB preserves intrinsic sensing characteristics of SPL, including photon sparsity, background contamination, and multi-return ambiguity, which collectively pose significant challenges for robust perception. While poses and depth references are derived from auxiliary LiDAR and cross-sensor calibration, minor SLAM drift and registration errors may introduce geometric uncertainties. Consequently, depth evaluation in STB should be interpreted relative to LiDAR-based references rather than absolute ground truth. The current release comprises 10 scenes and 10,297 views spanning diverse indoor and outdoor environments, collected under a unified SPL configuration to ensure consistent benchmarking conditions. Future extensions will expand scene diversity and explore heterogeneous sensing setups across different SPL devices and acquisition conditions. In addition, ambient illumination intensity is recorded during data capture. Although not directly incorporated into the current benchmark metrics, this metadata provides contextual information for analyzing photon-limited regimes and may support future research on illumination-aware reconstruction, sensing robustness, and adaptive perception under varying environmental conditions.


\bibliographystyle{splncs04}
\bibliography{main}

\end{document}


\title{\texorpdfstring{\datasetname: A Real-Captured Single Photon Perception Benchmark\\[0.5em]\large Supplementary Material}{\datasetname: A Real-Captured Single Photon Perception Benchmark - Supplementary Material}}

\author{Hongzhou Dong\inst{1} \and
Zili Zhang\inst{1} \and
Ziting Wen\inst{2} \and
Yiheng Qiang\inst{1} \and
Runrong Deng\inst{1} \and
Wenle Dong\inst{1} \and
Ziwen Jiang\inst{1} \and
Xinyang Li\inst{1} \and
Rui Lu\inst{1} \and
Shuoyao Sun\inst{1} \and
Wenyu Wang\inst{1} \and
Ziyi Xia\inst{1} \and
Haitao Zheng\inst{1} \and
Guodong Shi\inst{3} \and
Xiaoqiang Ren\inst{1}}

\authorrunning{H.~Dong et al.}

\institute{Shanghai University, Shanghai, China \and
Southern University of Science and Technology, Shenzhen, China \and
The University of Sydney, Sydney, Australia}

\maketitle

\appendix

\section{Hardware Specifications of the SPL Device}
\label{sec:supp_hardware_specs}

\begin{figure}[h]
    \centering
    \includegraphics[width=0.2\linewidth]{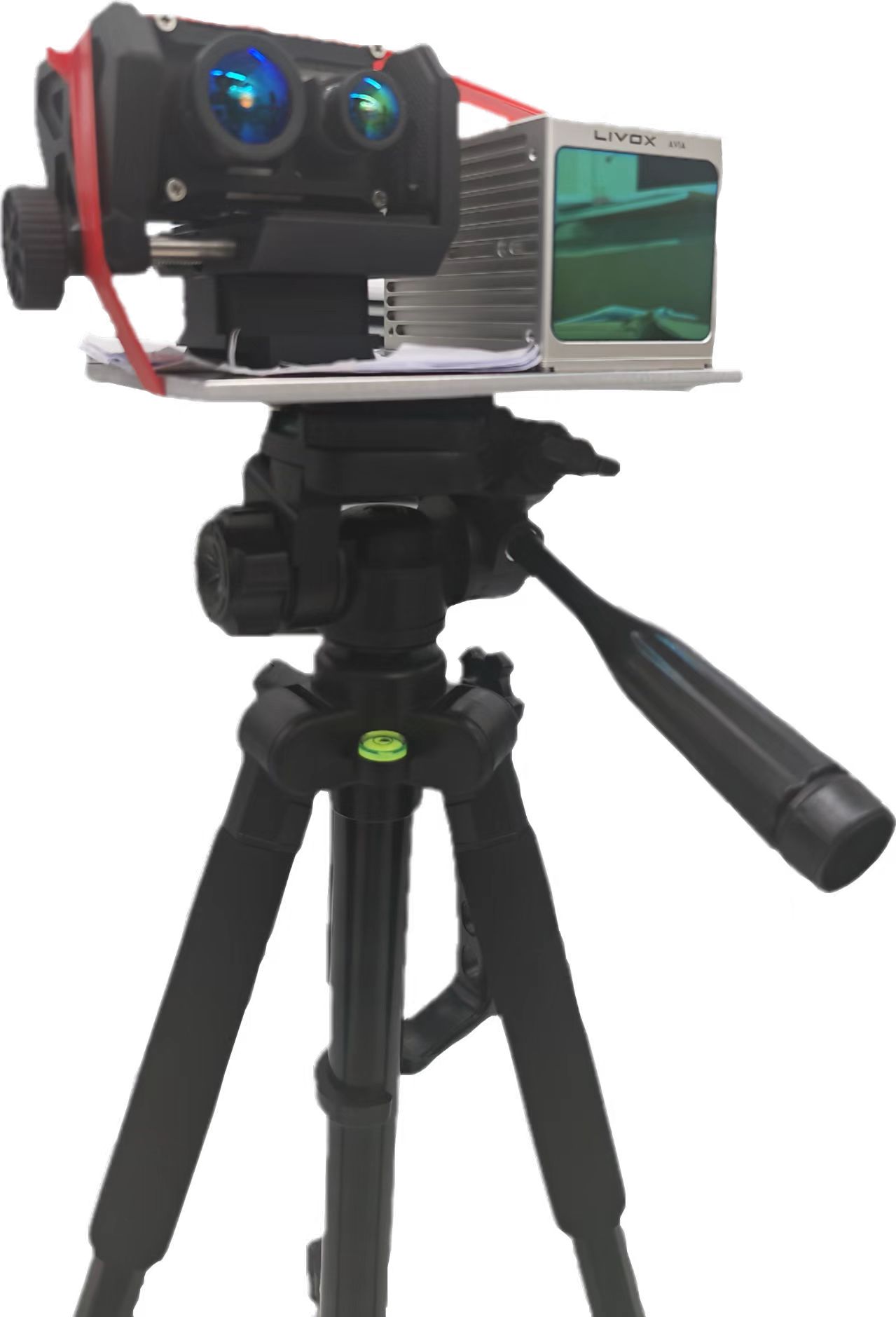}
    \caption{\textbf{Hardware setup for data collection.} The custom-designed sensor suite integrates the SPL device (left) and the Livox LiDAR (right) on a rigid mounting platform to ensure spatial calibration stability.}
    \label{fig:hardware_setup}
\end{figure}

In this section, we provide the detailed specifications and operating principles of the ADS6311 solid-state LiDAR used in \datasetname. The ADS6311 is a high-resolution Single Photon LiDAR (SPL) system characterized by its lack of mechanical moving parts, ensuring high reliability and a compact form factor~\cite{adaps_hawk}. The physical setup of our sensor suite is shown in Fig.~\ref{fig:hardware_setup}, where the SPL device (ADS6311) is positioned on the left and the Livox LiDAR on the right.

\subsection{Operating Principle}
The device operates on the principles of Direct Time of Flight (DToF) and Time-Correlated Single Photon Counting (TCSPC). The transmitter (Tx) utilizes a Vertical-Cavity Surface-Emitting Laser (VCSEL) array to emit ultrashort laser pulses (4--10 ns) at a wavelength of \(940~\mathrm{nm}\). These pulses are reflected by targets in the scene and captured by the receiver (Rx) optics. The receiver consists of a Single-Photon Avalanche Diode (SPAD) array synchronized with the Tx. By recording the precise time-stamps of photon detections over multiple emission cycles, the system constructs a histogram of photon counts across temporal bins, enabling precise distance measurement and transient imaging.

\subsection{Technical Specifications}
The ADS6311 provides a wide field of view (FOV) and high angular resolution, which are critical for the semantic understanding and reconstruction tasks in our benchmark. The raw SPAD resolution is \(768 \times 576\), which, after \(3 \times 3\) on-chip binning, yields a final depth map resolution of \(256 \times 192\). The key mechanical, electrical, and optical specifications are summarized in Table~\ref{tab:lidar_specs}.

\begin{table}[h]
\centering
\caption{\textbf{Key specifications of the ADS6311 solid-state LiDAR.}}
\label{tab:lidar_specs}
\small
\begin{tabular}{ll}
\toprule
\textbf{Parameter} & \textbf{Value} \\
\midrule
Scanning Method & Solid-state, electrical switching \\
Wavelength & \(940~\mathrm{nm}\) \\
Channels & \(32 \times 2\) \\
Depth Map Resolution & \(256 \times 192\) \\
Field of View & \(128^\circ \times 96^\circ\) \\
Angular Resolution & \(0.5^\circ \times 0.5^\circ\) \\
\midrule
Detection Range & Up to \(30~\mathrm{m}\) \\
Range Accuracy & \(< 5~\mathrm{cm}\) \\
Range Precision & \(< 1\%\) \\
Frame Rate & \(10-20~\mathrm{Hz}\) \\
Point Cloud Rate & \(983,040~\text{points/sec}\) \\

\bottomrule
\end{tabular}
\end{table}

\section{Calibration Detailes}
\label{sec:supp_a_calibration_reference}
\subsection{Intrinsic Parameters Calibration of the SPL Devices}

We calibrated the intrinsic parameters of the SPL devices using the MATLAB Camera Calibrator toolbox with a planar checkerboard target. 
The calibration board is a GP400-30-12$\times$9 checkerboard. 
For each device, calibration images were captured with the checkerboard placed at multiple positions, depths, and viewing angles to provide sufficient geometric constraints across the full field of view~\cite{calibration}.

Since our SPL system records transient histograms rather than standard RGB images, we first converted each measurement into a 2D intensity image by integrating photon counts over the temporal dimension. 
Checkerboard corners were detected on these intensity images and refined within the MATLAB Camera Calibrator pipeline. 
A standard pinhole camera model with lens distortion was adopted to estimate the intrinsic parameters.

The semantic understanding task involves two SPL devices, denoted as device 1 (p1) and device 2 (p2). 
In contrast, the depth estimation and multi-view reconstruction tasks only use data captured by device 2 (p2). 
Therefore, we report the calibrated intrinsics for both devices below, while noting that only the semantic track requires both sets of parameters.

The calibrated intrinsic matrix of device 1 (p1) is
\[
\mathbf{K}_{\mathrm{p1}} =
\begin{bmatrix}
120.9401 & 0 & 130.4126 \\
0 & 121.1257 & 97.1258 \\
0 & 0 & 1
\end{bmatrix},
\]
and its distortion coefficients are
\[
\mathbf{d}_{\mathrm{p1}} =
\left[
-0.2769,\ 
0.0623,\ 
0,\ 
0
\right].
\]

The calibrated intrinsic matrix of device 2 (p2) is
\[
\mathbf{K}_{\mathrm{p2}} =
\begin{bmatrix}
118.6515 & 0 & 130.6803 \\
0 & 118.7965 & 100.3606 \\
0 & 0 & 1
\end{bmatrix},
\]
and its distortion coefficients are
\[
\mathbf{d}_{\mathrm{p2}} =
\left[
-0.2769,\ 
0.0623,\ 
0,\ 
0
\right].
\]

Here, each distortion vector is organized as
\[
\mathbf{d} = [k_1,\, k_2,\, p_1,\, p_2],
\]
where \(k_1\) and \(k_2\) are the radial distortion coefficients, and \(p_1\) and \(p_2\) are the tangential distortion coefficients. 
For both devices, the estimated tangential distortion is zero.

All temporal bins of a given device share the same spatial intrinsics, since they are recorded on the same sensor plane. 
For reproducibility, the released dataset will include the calibrated intrinsic matrices and distortion coefficients for both devices.

\subsection{Extrinsic Parameters Calibration between the SPL sensor and the Livox LiDAR}
\label{sec:supp_calibration_details}

We additionally calibrate the extrinsic transformation between the SPL sensor and the Livox LiDAR using a checkerboard-based calibration procedure. Specifically, both sensors observe the same checkerboard target from multiple poses, and the relative pose between the two sensor coordinate systems is estimated from these paired observations. To reduce potential setup bias caused by sensor movement or mounting variations, we re-calibrate the SPL--Livox extrinsics before collecting each reconstruction sequence. The estimated extrinsic parameters are provided together with the corresponding reconstruction data for reproducibility.

\subsection{Instrument Response Function Characterization}
\label{sec:supp_irf}

We measured the instrument response function (IRF) of the SPL system using a
high-reflectance (99\%) ESR target under dark condition, with ambient illumination
close to $0$ lux~\cite{shin2016photon}. The target was placed at three fixed distances, i.e.,
$28$\,cm, $54$\,cm, and $96$\,cm.

\begin{figure}[t]
    \centering
    \includegraphics[width=\linewidth]{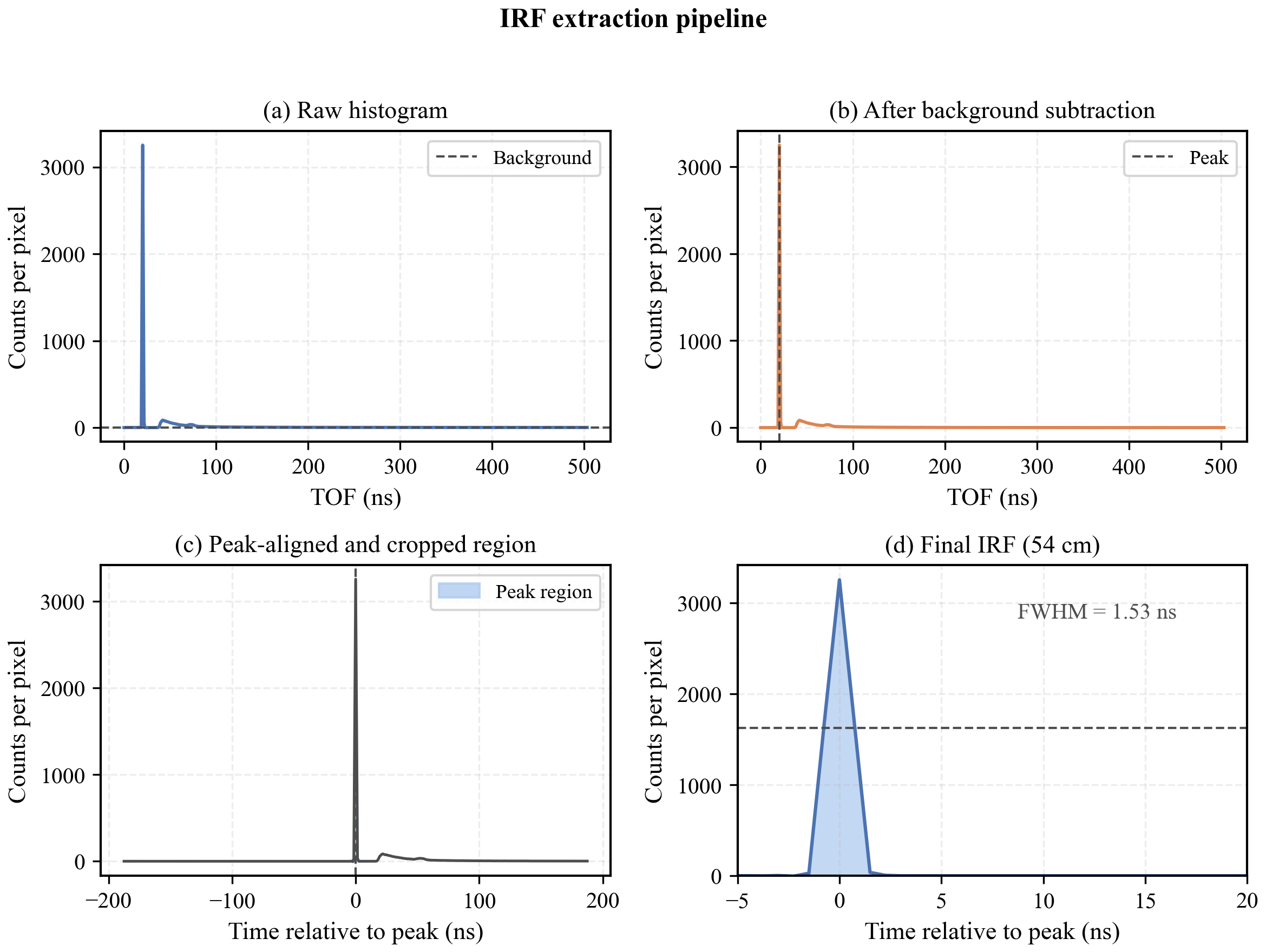}
    \caption{
    \textbf{IRF extraction pipeline.}
    From left to right and top to bottom: ROI-averaged raw histogram, background-subtracted
    histogram, peak-aligned and cropped pulse region, and the final extracted IRF with
    the FWHM indicated.
    }
    \label{fig:irf_pipeline}
\end{figure}

For each measurement, we first formed an intensity image by taking the maximum
value along the temporal dimension, and then automatically extracted a target
ROI by Thresholdinging and morphological Matched Filtering. The ROI-averaged histogram was
used for IRF extraction. We estimated the background level from the pre-peak bins,
subtracted it from the histogram, aligned the dominant peak with sub-bin precision,
and cropped a local window around the peak. The right tail was further truncated
once the response stayed below a small fraction of the peak for a consecutive
interval. The resulting signal was taken as the extracted IRF. The overall
extraction process is illustrated in Fig.~\ref{fig:irf_pipeline}.

\begin{figure}[t]
    \centering
    \includegraphics[width=0.75\linewidth]{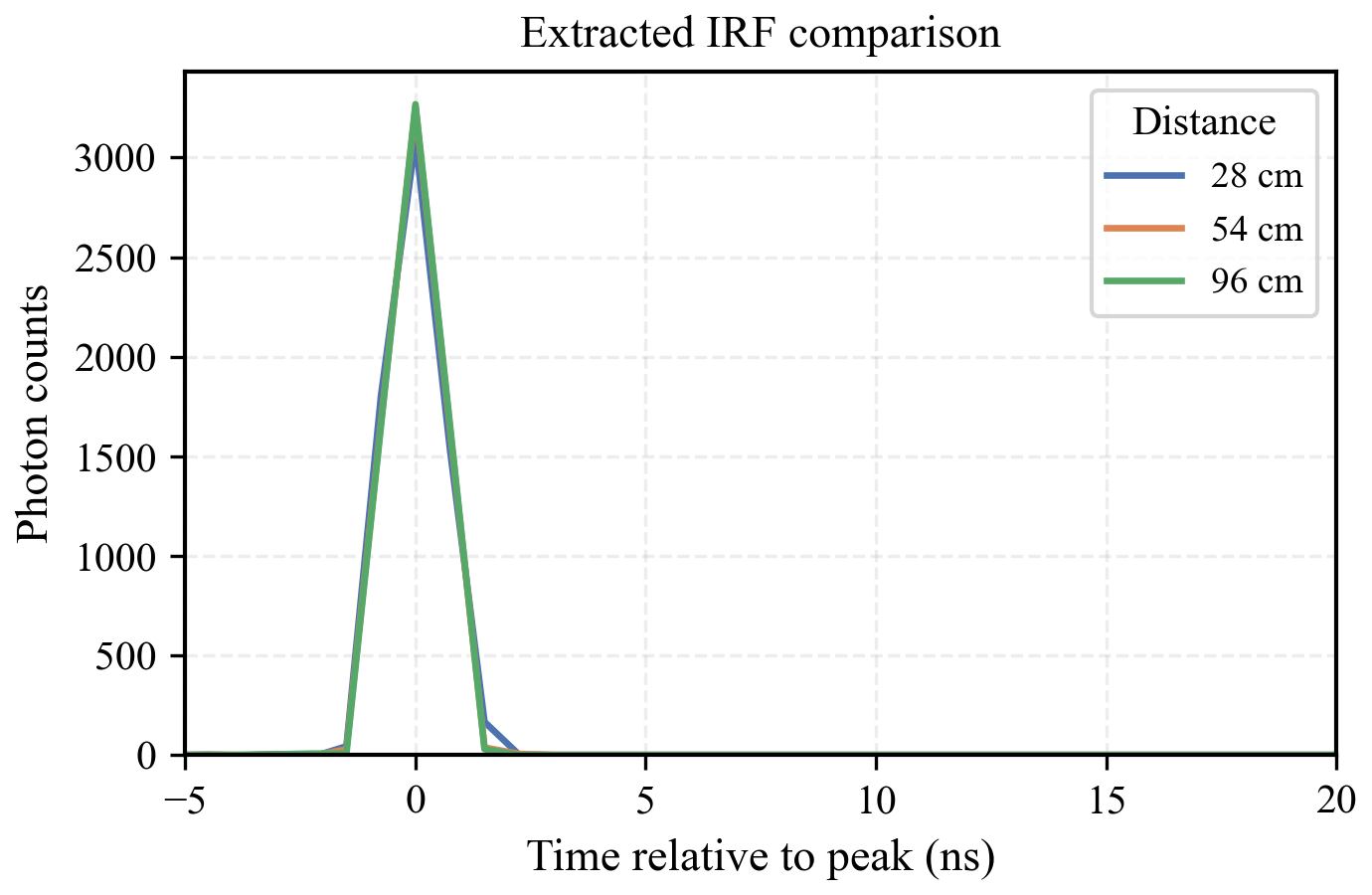}
    \caption{
    \textbf{Comparison of extracted IRFs at different distances.}
    Extracted IRFs measured at $28$\,cm, $54$\,cm, and $96$\,cm under dark condition
    (ambient illumination close to $0$ lux). The histograms are aligned to their
    dominant peaks to facilitate comparison of temporal width and trailing response.
    }
    \label{fig:irf_compare}
\end{figure}

We report the full width at half maximum (FWHM) as the main summary of temporal
pulse width. Fig.~\ref{fig:irf_compare} compares the extracted IRFs at different
distances, showing that the main pulse shape remains largely consistent while
exhibiting slight variations in temporal spread and trailing response.

\section{Detailed Experimental Results}
\label{sec:supp_b_depth}

\subsection{Detailed Results for Depth Estimation}
\label{sec:supp_depth_estimation}

\begin{figure*}[t]
    \centering
    \includegraphics[width=0.9\textwidth]{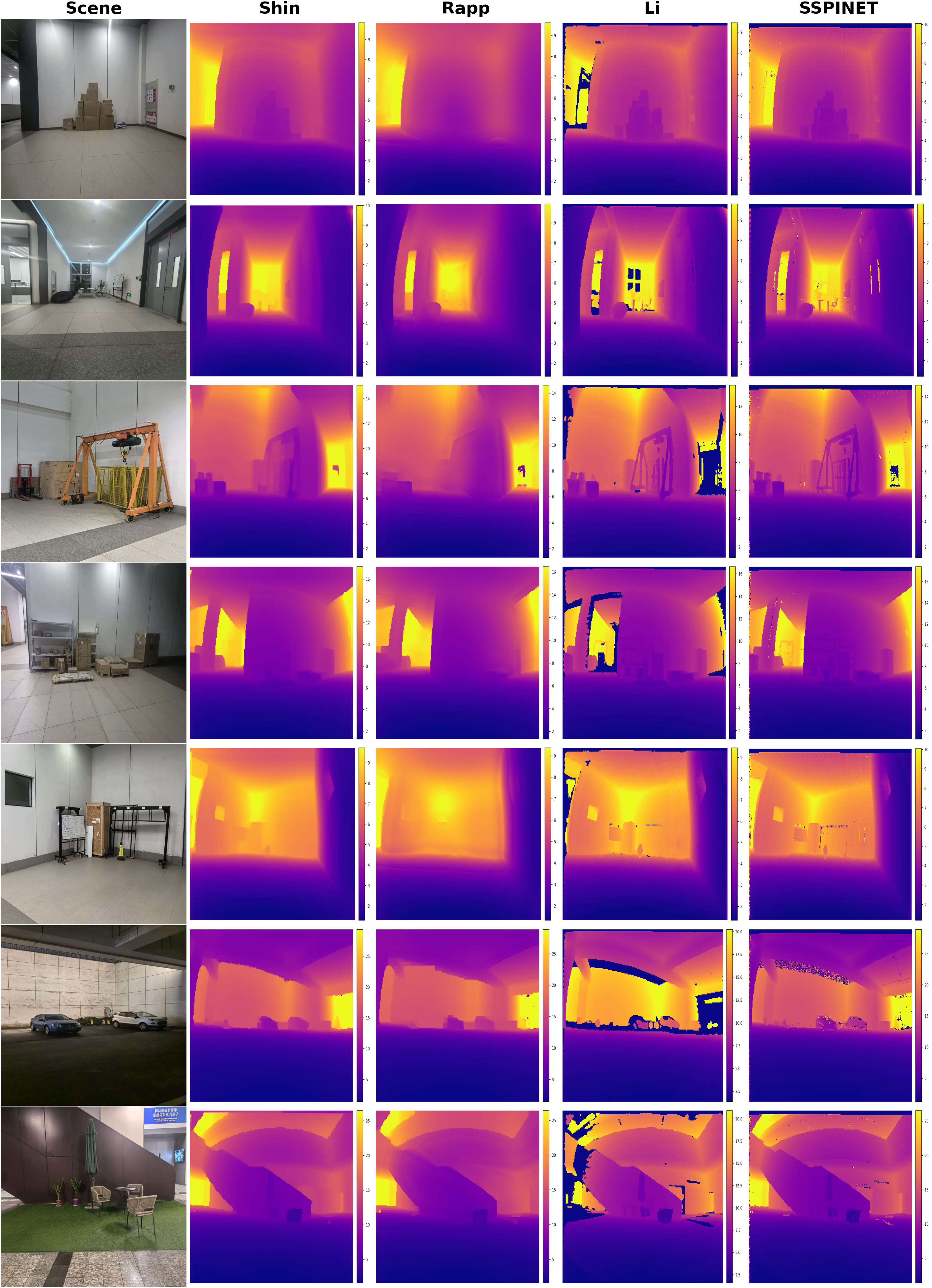}
   \caption{\textbf{Qualitative comparison of depth estimation results on our dataset.}
From left to right are the input scene image and the depth predictions produced by Shin, Rapp, Li, and SSPINET, respectively.}
    \label{fig:supp_depth_compare}
\end{figure*}

Additional visual comparisons for depth estimation is provided in Fig.~\ref{fig:supp_depth_compare}. 
The examples cover a variety of indoor scenes with different layouts and object configurations.

\subsection{Detailed Results for Multi-Vie 3D Reconstruction}
\label{sec:supp_c_reconstruction}

Following the same protocol as in the main paper, we further evaluate multi-view reconstruction under the 3-view and 5-view settings, while also including the 10-view results reported in the main paper for completeness. We consider the same transient-aware baselines, \emph{TransientNeRF}~\cite{Malik2023TransientNeRF} and \emph{Transientangelo}~\cite{2024Transientangelo}, and keep all training and evaluation settings unchanged. We report novel-view rendering metrics for intensity (SSIM~\cite{wang2004image}, LPIPS~\cite{zhang2018unreasonable}), depth ($L_1$), and histogram (PSNR) outputs.

In addition, we include several non-transient SPL reconstruction baselines for comparison. Unlike transient-aware methods, these approaches operate only on depth and intensity images extracted from the raw single-photon measurements, rather than directly modeling the transient histograms. As a result, part of the temporal information is discarded before reconstruction, leading to a larger information gap, especially under sparse-view or low-photon conditions. Among the selected baselines, TSDF Fusion~\cite{curless1996volumetric} and Reprojection are representative classical pipelines, while FrugalNeRF~\cite{lin2025frugalnerf} is a recent few-shot learning based reconstruction method.

\begin{figure*}[t]
    \centering
    \includegraphics[width=\textwidth]{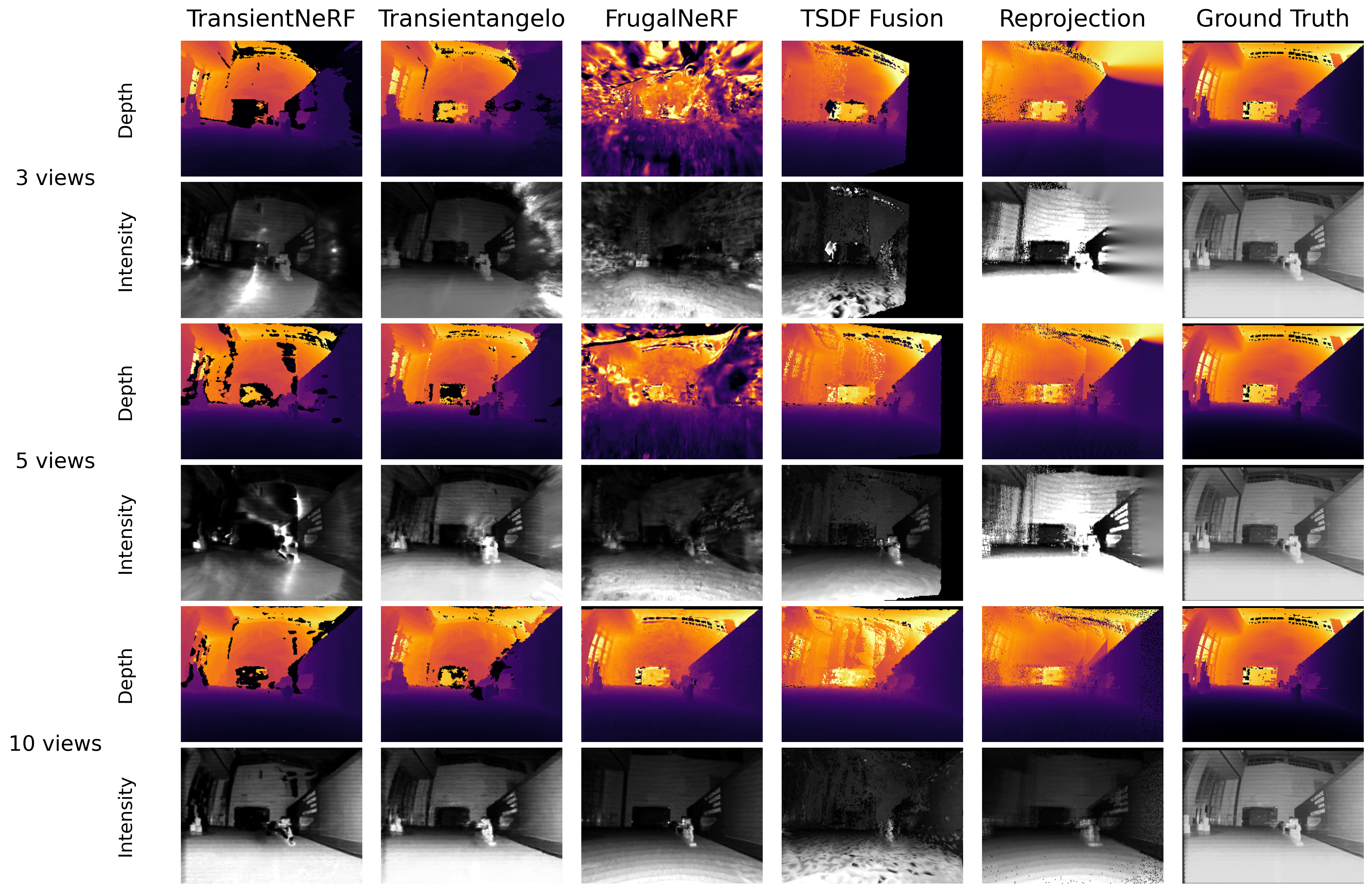}
    \caption{\textbf{Qualitative comparison of reconstruction results under different input-view settings on \datasetabbr.}
    We visualize novel-view depth and intensity predictions produced by different methods under 3-view, 5-view, and 10-view settings. 
    Each view setting contains two rows corresponding to depth and intensity renderings, respectively.}
    \label{fig:supp_recon_compare_views}
\end{figure*}

Table~\ref{tab:supp_recon_all_views} summarizes all additional reconstruction results. Overall, transient-aware methods consistently achieve stronger appearance and geometry reconstruction quality than non-transient baselines. While increasing the number of input views can improve some metrics, the gains are generally limited for non-transient methods and are not always monotonic. These results further support the importance of explicitly leveraging transient histogram measurements, rather than relying only on derived intensity and depth representations.To complement these metrics, Fig.~\ref{fig:supp_recon_compare_views} presents qualitative comparisons of the corresponding novel-view depth and intensity reconstructions.

\begin{table*}[t]
\centering
\caption{\textbf{Additional reconstruction results on \datasetabbr\ under 3-view, 5-view, and 10-view settings.}
We report novel-view rendering metrics for intensity (SSIM, LPIPS), depth ($L_1$), and histogram (PSNR) outputs. Histogram-domain PSNR is only reported for transient-aware methods, since non-transient baselines do not explicitly model transient histograms. The 10-view results are copied from the main paper for reference.}
\label{tab:supp_recon_all_views}
\setlength{\tabcolsep}{3.5pt}
\small
\resizebox{\textwidth}{!}{
\begin{tabular}{c cccc cccc cccc}
\toprule
\multirow{3}{*}{Method}
& \multicolumn{4}{c}{3-view}
& \multicolumn{4}{c}{5-view}
& \multicolumn{4}{c}{10-view} \\
\cmidrule(lr){2-5}\cmidrule(lr){6-9}\cmidrule(lr){10-13}
& \multicolumn{2}{c}{Intensity} & Depth & Histogram
& \multicolumn{2}{c}{Intensity} & Depth & Histogram
& \multicolumn{2}{c}{Intensity} & Depth & Histogram \\
\cmidrule(lr){2-3}\cmidrule(lr){4-4}\cmidrule(lr){5-5}
\cmidrule(lr){6-7}\cmidrule(lr){8-8}\cmidrule(lr){9-9}
\cmidrule(lr){10-11}\cmidrule(lr){12-12}\cmidrule(lr){13-13}
& SSIM $\uparrow$ & LPIPS $\downarrow$ & $L_1$ $\downarrow$ & PSNR $\uparrow$
& SSIM $\uparrow$ & LPIPS $\downarrow$ & $L_1$ $\downarrow$ & PSNR $\uparrow$
& SSIM $\uparrow$ & LPIPS $\downarrow$ & $L_1$ $\downarrow$ & PSNR $\uparrow$ \\
\midrule
\multicolumn{13}{c}{\textit{Transient-aware methods}} \\
\midrule
TransientNeRF~\cite{Malik2023TransientNeRF}
& 0.7836 & 0.2323 & 0.9518 & 45.4687
& 0.8046 & 0.2272 & 0.8773 & 47.5624
& 0.7717 & 0.2315 & 0.7836 & 45.4251 \\

Transientangelo~\cite{2024Transientangelo}
& 0.7260 & 0.2837 & 1.6516 & 42.0865
& 0.7065 & 0.2917 & 1.6516 & 42.1014
& 0.7597 & 0.2501 & 0.8710 & 41.5735 \\
\midrule
\multicolumn{13}{c}{\textit{Non-transient methods}} \\
\midrule
TSDF Fusion~\cite{curless1996volumetric}
& 0.5337 & 0.3361 & 2.6614 & --
& 0.4279 & 0.4539 & 2.6459 & --
& 0.4619 & 0.4153 & 2.7361 & -- \\

Reprojection
& 0.6235 & 0.2910 & 2.8886 & --
& 0.5183 & 0.5212 & 2.7369 & --
& 0.5617 & 0.5438 & 2.6003 & -- \\

FrugalNeRF~\cite{lin2025frugalnerf}
& 0.5479 & 0.3552 & 4.3714 & --
& 0.5789 & 0.3303 & 4.0621 & --
& 0.6155 & 0.3105 & 3.7966 & -- \\
\bottomrule
\end{tabular}
}
\end{table*}

\begin{figure*}[t]
    \centering
    \begin{subfigure}{0.9\textwidth}
        \centering
        \includegraphics[width=\textwidth]{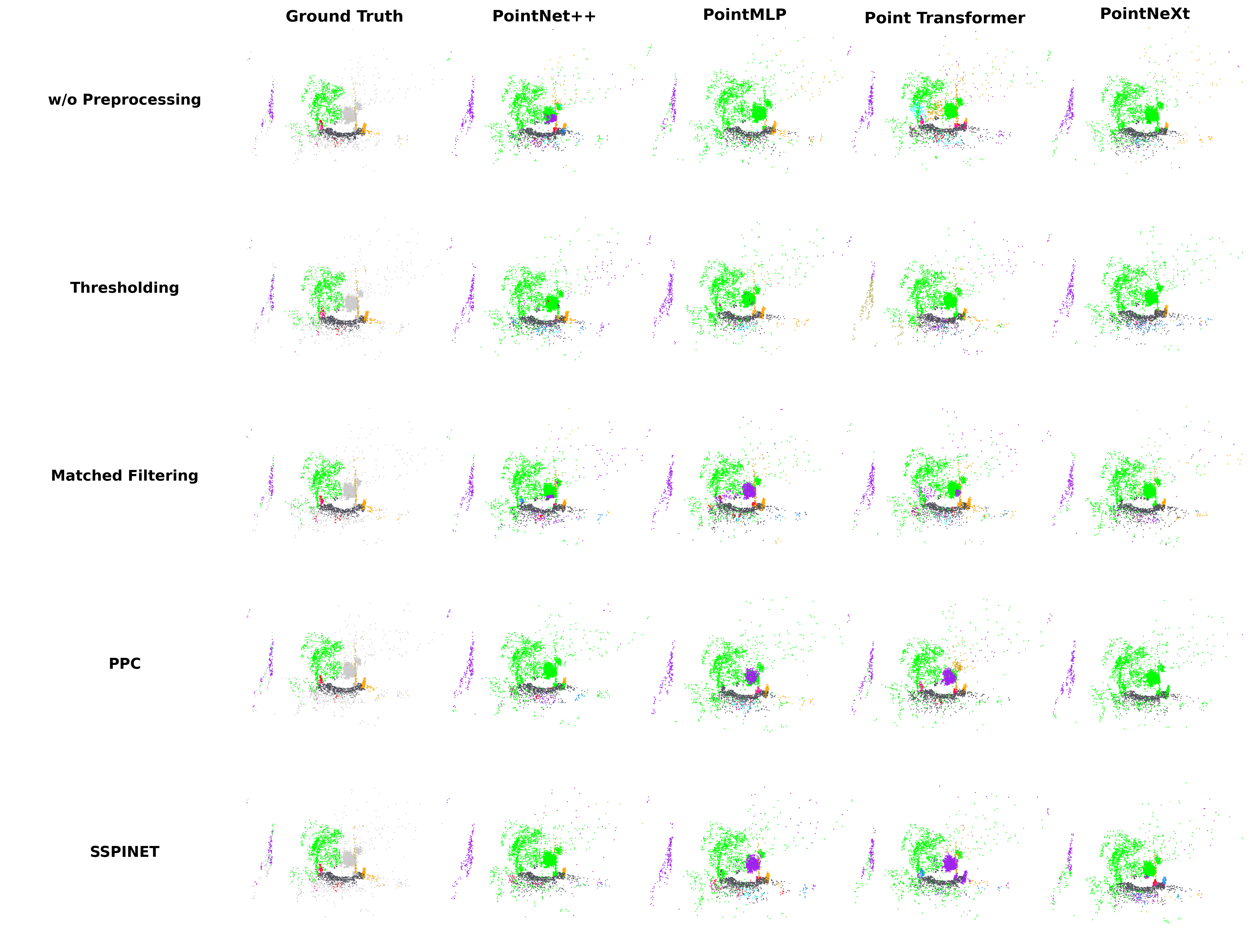}
    \end{subfigure}
    
    \vspace{0.5em}
    
    \begin{subfigure}{0.9\textwidth}
        \centering
        \includegraphics[width=\textwidth]{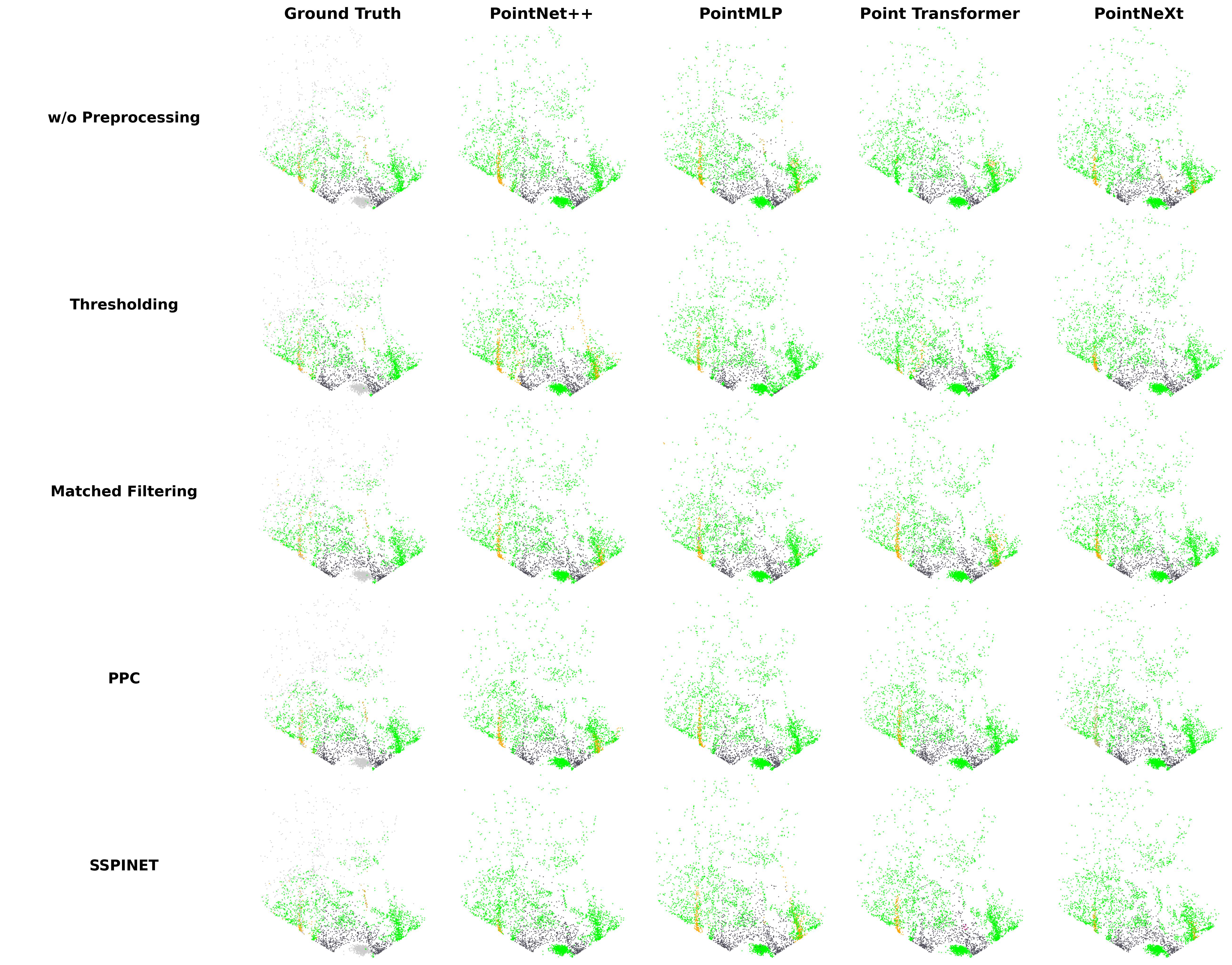}
    \end{subfigure}
    \caption{\textbf{Qualitative results for 3D semantic segmentation on our dataset.} 
We present two representative examples from the dataset. Columns correspond to the Ground Truth and the predictions obtained with different preprocessing and backbone combinations.}
 \label{fig:supp_semantic_compare}
\end{figure*}

\subsection{Detailed Results for Semantic Understanding}
For completeness, we provide the detailed semantic understanding results on \datasetabbr\ in Tables~\ref{tab:supp_semantic_detailed_merged} and the qualitative examples in Fig.~\ref{fig:supp_semantic_compare}. Specifically, we report the per-class IoU (\%) and mIoU (\%) of different segmentation backbones under different preprocessing methods, while Fig.~\ref{fig:supp_semantic_compare} shows two representative samples from the dataset.

\begin{table*}[t]
\caption{\textbf{Detailed semantic understanding results on \datasetabbr.}
We report per-class IoU (\%) for different segmentation backbones under different preprocessing methods, together with mIoU (\%).}
\label{tab:supp_semantic_detailed_merged}
\setlength{\tabcolsep}{4pt}
\renewcommand{\arraystretch}{1.02}
\footnotesize
\resizebox{\textwidth}{!}{%
\begin{tabular}{ll|cccccccccccccc}
\toprule
Methods & Preproc. 
& Tree & Road & Fence & Person & Non-motor
& Car & St.~Light & Signage & T.~Light & Door
& Building & Wall & In.~Roof & mIoU \\
\midrule
\multirow{5}{*}{PointMLP~\cite{ma2022rethinking}}
& w/o Preprocessing                & 86.51 & 90.69 & 58.34 & 23.04 & 31.00 & 38.03 & 29.03 & 19.74 & 6.11 & 28.63 & 71.16 & 66.74 & 89.70 & 49.13 \\
& Thresholding                     & 87.13 & 91.64 & 62.94 & 31.63 & 41.56 & 39.31 & 28.20 & 26.32 & 3.17 & 29.47 & 69.69 & 73.61 & 90.89 & 51.97 \\
& Matched Filtering~\cite{turin1960matched}& 87.73 & 90.48 & 58.99 & 18.08 & 37.90 & 42.99 & 34.14 & 18.37 & 2.50 & 29.73 & 68.55 & 69.11 & 91.08 & 49.97 \\
& PPC~\cite{Goyal_2025_ICCV}       & 86.49 & 92.41 & 62.82 & 28.56 & 42.04 & 46.67 & 29.89 & 28.45 & 2.68 & 17.39 & 73.61 & 73.38 & 91.81 & 52.02 \\
& SSPINET~\cite{yao2022dynamic}    & 88.49 & 91.66 & 61.42 & 29.38 & 41.77 & 41.21 & 32.44 & 22.59 & 1.77 & 19.42 & 71.84 & 69.50 & 89.81 & 50.87 \\
\midrule
\multirow{5}{*}{PointNet++~\cite{qi2017pointnet++}}
& w/o Preprocessing                & 86.61 & 91.39 & 56.24 & 26.10 & 37.54 & 38.84 & 31.15 & 23.44 & 0.00 & 6.67  & 68.21 & 65.24 & 89.18 & 47.74 \\
& Thresholding                     & 86.37 & 92.29 & 59.21 & 33.43 & 37.39 & 38.68 & 31.08 & 23.81 & 0.00 & 21.90 & 64.22 & 71.45 & 89.63 & 49.96 \\
& Matched Filtering~\cite{turin1960matched}& 87.68 & 91.55 & 58.16 & 26.75 & 36.79 & 39.31 & 32.46 & 22.84 & 0.00 & 12.72 & 71.79 & 66.72 & 90.22 & 49.00 \\
& PPC~\cite{Goyal_2025_ICCV}       & 86.03 & 92.62 & 61.45 & 26.07 & 43.40 & 43.79 & 31.93 & 26.33 & 0.00 & 14.28 & 69.05 & 71.39 & 87.66 & 50.31 \\
& SSPINET~\cite{yao2022dynamic}    & 87.48 & 91.98 & 60.18 & 26.99 & 41.53 & 43.06 & 31.46 & 22.55 & 0.00 & 15.94 & 71.91 & 69.51 & 90.59 & 50.25 \\
\midrule
\multirow{5}{*}{PointNeXt~\cite{qian2022pointnext}}
& w/o Preprocessing                & 85.69 & 90.52 & 51.14 & 5.90  & 22.54 & 24.91 & 23.79 & 2.74  & 0.00 & 0.00  & 68.60 & 65.64 & 89.05 & 40.81 \\
& Thresholding                     & 84.72 & 91.42 & 52.37 & 26.28 & 33.30 & 21.52 & 23.56 & 10.86 & 0.00 & 0.00  & 63.56 & 67.36 & 90.28 & 43.48 \\
& Matched Filtering~\cite{turin1960matched}& 84.18 & 90.18 & 48.37 & 3.04  & 22.24 & 25.81 & 21.91 & 6.50  & 0.00 & 0.00  & 68.15 & 65.12 & 88.87 & 40.34 \\
& PPC~\cite{Goyal_2025_ICCV}       & 81.84 & 91.44 & 46.19 & 14.06 & 27.62 & 25.96 & 24.56 & 0.71  & 0.00 & 0.00  & 65.82 & 61.15 & 84.42 & 40.29 \\
& SSPINET~\cite{yao2022dynamic}    & 86.19 & 90.88 & 52.11 & 12.28 & 34.97 & 30.15 & 24.18 & 6.74  & 0.00 & 0.00  & 69.91 & 64.39 & 90.87 & 43.28 \\
\midrule
\multirow{5}{*}{Point Transformer~\cite{zhao2021point}}
& w/o Preprocessing                & 87.39 & 91.62 & 60.65 & 17.11 & 39.56 & 37.05 & 30.47 & 20.29 & 0.36 & 28.11 & 71.04 & 69.01 & 91.05 & 49.52 \\
& Thresholding                     & 86.58 & 91.55 & 58.97 & 26.09 & 37.59 & 32.82 & 30.14 & 27.28 & 5.73 & 31.11 & 67.07 & 69.06 & 90.33 & 50.33 \\
& Matched Filtering~\cite{turin1960matched}& 87.96 & 91.54 & 58.56 & 25.89 & 42.31 & 38.47 & 30.08 & 25.98 & 0.00 & 23.54 & 72.71 & 67.22 & 90.10 & 50.34 \\
& PPC~\cite{Goyal_2025_ICCV}       & 86.31 & 92.68 & 61.19 & 25.63 & 41.62 & 39.76 & 27.99 & 22.25 & 0.43 & 14.90 & 72.40 & 72.30 & 91.11 & 49.89 \\
& SSPINET~\cite{yao2022dynamic}    & 87.75 & 91.83 & 59.61 & 22.83 & 44.25 & 41.22 & 30.86 & 21.00 & 3.00 & 19.42 & 70.10 & 68.14 & 88.87 & 49.91 \\
\bottomrule
\end{tabular}%
}
\end{table*}

\begin{figure*}[t]
    \centering

    \begin{subfigure}[t]{0.9\textwidth}
        \centering
        \includegraphics[width=\textwidth]{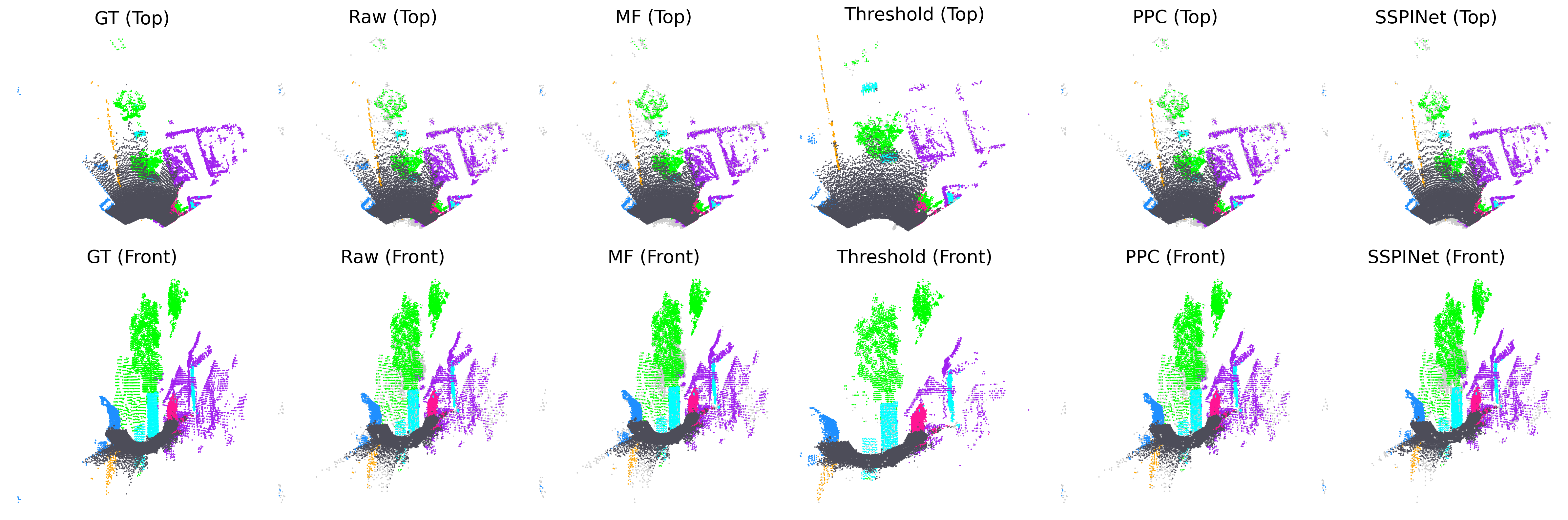}
        \caption{Outdoor street scene.}
    \end{subfigure}

    \vspace{6pt}

    \begin{subfigure}[t]{0.9\textwidth}
        \centering
        \includegraphics[width=\textwidth]{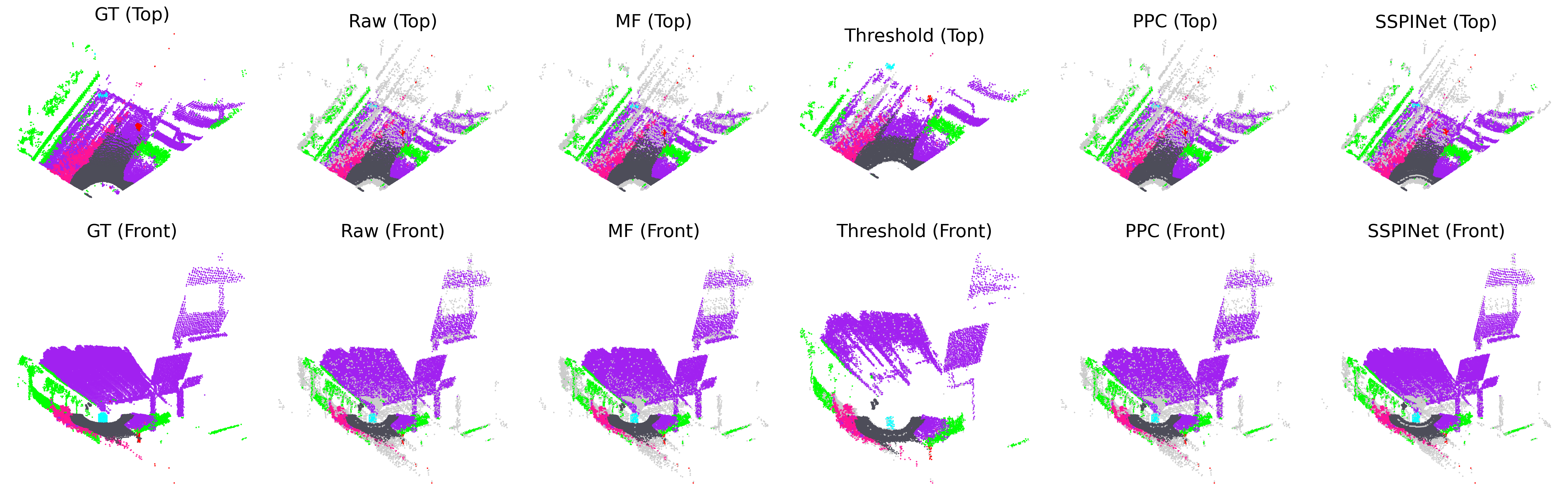}
        \caption{Outdoor street scene.}
    \end{subfigure}

    \vspace{6pt}

    \begin{subfigure}[t]{0.9\textwidth}
        \centering
        \includegraphics[width=\textwidth]{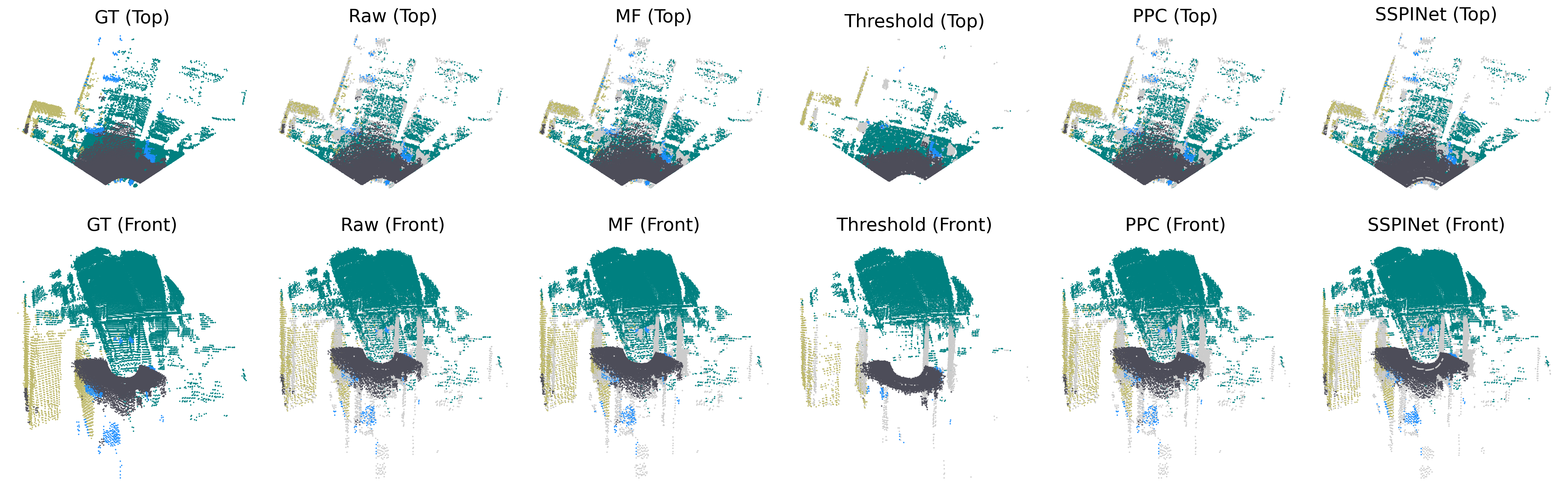}
        \caption{Indoor parking-lot scene.}
    \end{subfigure}

    \vspace{6pt}

    \begin{subfigure}[t]{0.9\textwidth}
        \centering
        \includegraphics[width=\textwidth]{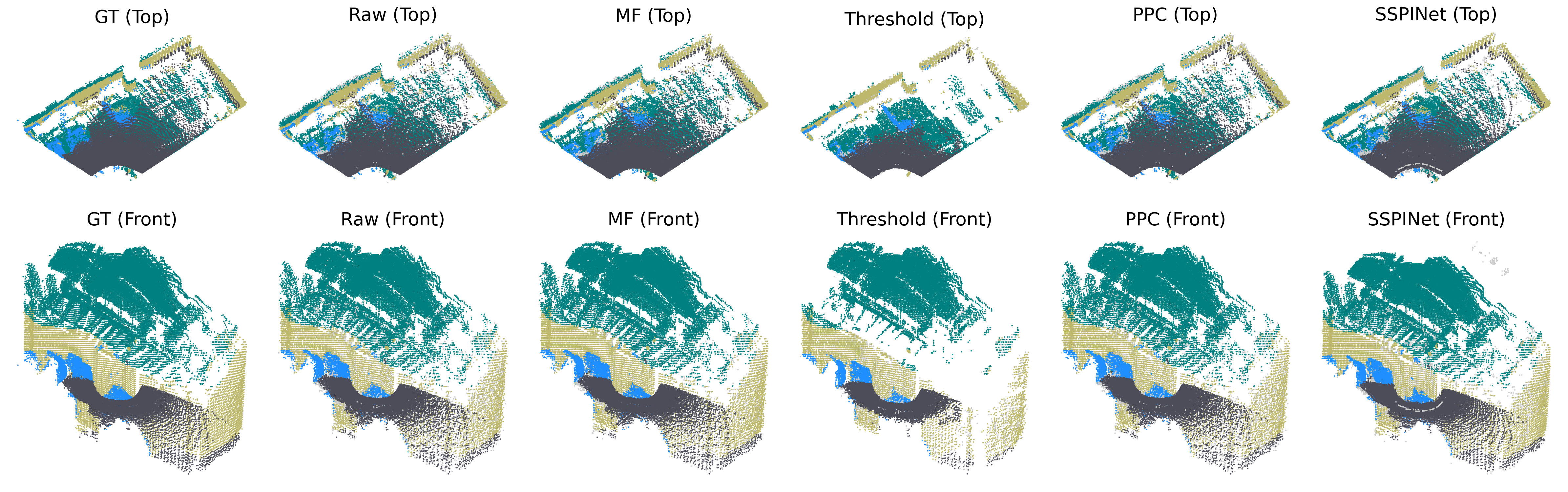}
        \caption{Indoor parking-lot scene.}
    \end{subfigure}

    \caption{\textbf{Qualitative visualization of different preprocessing methods on representative scenes.}
    (a) and (b) show two outdoor street scenes, while (c) and (d) show two indoor parking-lot scenes. For each subfigure, the upper row presents the top view and the lower row presents the front view. The columns show GT, Raw, Matched Filtering(MF)~\cite{turin1960matched} , Thresholding, PPC~\cite{Goyal_2025_ICCV}, and SSPINET~\cite{yao2022dynamic}, respectively.}
    \label{fig:supp_preproc_vis}
\end{figure*}

\subsection{Qualitative Visualization of Preprocessing Results}
In addition to the quantitative results, we further provide qualitative visualizations of representative indoor and outdoor scenes after different preprocessing methods. Specifically, Fig.~\ref{fig:supp_preproc_vis} presents four typical examples: (a) and (b) correspond to outdoor street scenes, while (c) and (d) correspond to indoor parking-lot scenes. For each example, the upper row shows the top view and the lower row shows the front view. From left to right, the columns correspond to GT, Raw, MF, Thresholding, PPC~\cite{Goyal_2025_ICCV}, and SSPINET~\cite{yao2022dynamic}.For completeness, we also provide qualitative visualizations of representative indoor and outdoor scenes under different preprocessing methods in Fig.~\ref{fig:supp_preproc_vis}. Specifically, (a) and (b) show outdoor street scenes, while (c) and (d) show indoor parking-lot scenes. For each example, the upper row presents the top view and the lower row presents the front view. The columns correspond to GT, Raw, Matched Filtering(MF)~\cite{turin1960matched}, Thresholding, PPC~\cite{Goyal_2025_ICCV}, and SSPINET~\cite{yao2022dynamic}, respectively.

\section{Split Details and Release Protocol}
\subsection{Split Details}
For the Semantic Understanding task, the \datasetname\ (\datasetabbr) contains a total of \textbf{10,297} frames (views) captured across \textbf{27} unique sequences. The dataset is partitioned into a training set of \textbf{8,297} samples and a testing set of \textbf{2,000} samples using a frame-level random split to ensure that both sets represent the full diversity of lighting conditions and scene geometries. The specific partition will be provided along with the data.

\subsection{Release Protocol}
We are committed to the principles of open science. The dataset and associated resources will be released as follows:
\begin{itemize}[noitemsep]
    \item \textbf{License:} The dataset is released under the Creative Commons Attribution-NonCommercial-ShareAlike 4.0 International (CC BY-NC-SA 4.0) license.
    \item \textbf{Codebase:} The complete data,including raw histogram files and \texttt{.npy} annotation files, camera poses ,will be released upon acceptance.
\end{itemize}

\bibliographystyle{splncs04}
\bibliography{main}